\documentclass[11pt]{article}

\usepackage[T1]{fontenc}

\usepackage{helvet}
\usepackage[margin=1in]{geometry}
\usepackage{titlesec}
\usepackage{graphicx}
\usepackage{parskip}
\usepackage[dvipsnames]{xcolor}
\usepackage{tcolorbox}
\usepackage[numbers,square]{natbib}  
\usepackage{amsmath, amssymb}
\usepackage{fancyhdr}
\usepackage{microtype}
\usepackage{enumitem}
\usepackage[hidelinks]{hyperref}  
\usepackage[left]{lineno}

\usepackage{url}
\usepackage{booktabs}
\usepackage{amsfonts}
\usepackage{nicefrac}
\usepackage{float}
\usepackage{algorithm}
\usepackage{algpseudocode}
\usepackage{inconsolata}
\usepackage[labelfont=bf]{caption}
\usepackage{multirow}
\usepackage{refcount}
\usepackage{footmisc}
\usepackage{wrapfig}

\definecolor{metaBlue}{RGB}{24,119,242}
\definecolor{abstractBoxBG}{RGB}{245,245,250}

\titleformat{\section}{\color{black}\sffamily\Large\bfseries}{}{0pt}{}
\titlespacing*{\section}{0pt}{1.5\baselineskip}{1\baselineskip}

\titleformat{\subsection}{\color{black}\sffamily\large\bfseries}{}{0pt}{}
\titlespacing*{\subsection}{0pt}{1.5\baselineskip}{1\baselineskip}

\usepackage{xspace}
\newcommand{\scititle}{Zero-shot World Models Are Developmentally Efficient Learners}

\begin{document}
\thispagestyle{empty}

\newcommand{\changes}[1]{{#1}}

\vspace*{-2em}
\begin{tcolorbox}[
  colback=abstractBoxBG,
  colframe=abstractBoxBG,  
  boxrule=0pt,
  arc=4pt,
  left=12pt, right=12pt, top=10pt, bottom=12pt,
  width=\textwidth,
  enlarge left by=0mm,
  enlarge right by=0mm
]

{\sffamily\LARGE\bfseries \scititle \\[0.5em]}
{\sffamily\textbf{Khai Loong Aw}, \textbf{Klemen Kotar}, \textbf{Wanhee Lee}, \textbf{Seungwoo Kim}, \textbf{Khaled Jedoui}, \\
\textbf{Rahul Venkatesh}, \textbf{Lilian Naing Chen}, \textbf{Michael C. Frank}, \textbf{Daniel L.K. Yamins}}\\[0.3em]
\small Stanford University\\
Corresponding author: khaiaw@stanford.edu \\
[1em]
\textbf{Abstract.} Young children demonstrate early abilities to understand their physical world, estimating depth, motion, object coherence, interactions, and many other aspects of physical scene understanding. Children are both data-efficient and flexible cognitive systems, creating competence despite extremely limited training data, while generalizing to myriad untrained tasks -- a major challenge even for today's best AI systems. Here we introduce a novel computational hypothesis for these abilities, the Zero-shot World Model (ZWM). ZWM is based on three principles: a sparse temporally-factored predictor that decouples appearance from dynamics; zero-shot estimation through approximate causal inference; and composition of inferences to build more complex abilities. We show that ZWM can be learned from the first-person experience of a single child, rapidly generating competence across multiple physical understanding benchmarks. \changes{It also shows progressive, staged emergence of capacities during learning and builds brain-like internal representations.} Our work presents a blueprint for efficient and flexible learning from human-scale data, advancing both a computational account of children's early physical understanding and a path toward data-efficient AI systems.


\end{tcolorbox}

\vspace{-0.5em}

\begin{figure}[H]
	\centering
	\includegraphics[width=1.0\textwidth]{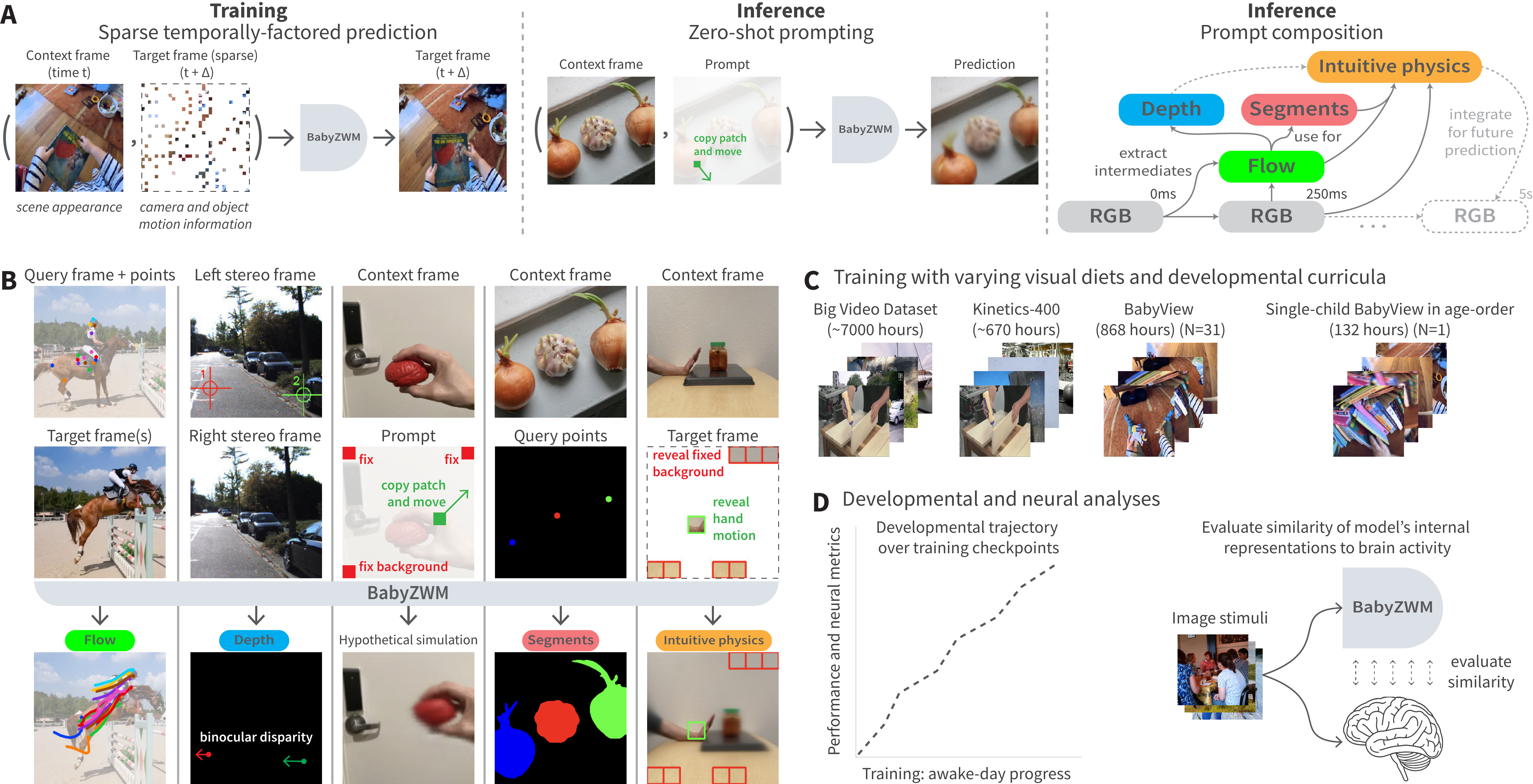}
	\caption{
        \textbf{Overview.}
        (\textbf{A}) The Zero-shot World Model (ZWM) framework has three design principles: temporally-factored prediction to flexibly separate appearance from dynamics; zero-shot extraction of visual-cognitive structures from the predictor through approximate causal inference; and composing extractors together to achieve increasingly complex inference abilities.
        (\textbf{B}) After self-supervised pretraining, ZWM can perform diverse visual-cognitive tasks zero-shot, i.e., without any additional training or examples.
        (\textbf{C}) We train ZWM with varying visual diets and single-child developmental curricula.
        (\textbf{D}) We evaluate BabyZWM's task performance across training checkpoints (developmental trajectory) and the similarity of its internal representations with brain responses.
    }
	\label{fig:main_framework}
\end{figure}

\noindent
By early childhood, humans demonstrate rich visual-cognitive abilities, exhibiting strong capacities on a diverse set of physical understanding tasks \cite{kellman_perception_1983, baillargeon_object_1985, spelke_origins_1992, spelke_core_2000, carey_origin_2009, spelke_what_2022}. The mechanisms behind these abilities are remarkably powerful, in two related but distinct senses.  First, they are \emph{data-efficient}, in that children demonstrate these capacities
despite highly limited ``training data'' afforded by the first-person experience of a single individual.  Second, they are \emph{flexible}, supporting performance of new task abilities from an existing general-purpose representation without task-specific examples (``zero-shot''), e.g., tracking motion, estimating depth, and intuitive physics.  Motivated by the sophistication and early emergence of infants' object and physical knowledge, some researchers have argued that babies have innate biases for visual cognition \cite{spelke_origins_1992, spelke_core_2000, carey_origin_2009, spelke_what_2022}. 
However, ``innateness'' can mean different things, e.g., the \emph{learning machinery} (objectives, architectures, programs) or \emph{content} (representational primitives and concepts). Here we ask: what ingredients are required to achieve data-efficient and flexible (zero-shot) visual cognition from early experience?

We approach this question using computational models of visual learning and development. Inspired by neurophysiological observations \cite{fukushima_neocognitron_1980, lecun_backpropagation_1989}, deep neural networks (DNNs) have emerged as the most task-performant models for visual tasks as well as the most accurate models of neural responses across the visual cortex \cite{yamins_performance-optimized_2014, guclu_deep_2015, yamins_using_2016, khaligh-razavi_deep_2014, cadena_deep_2019} and for human-like error patterns \cite{rajalingham_large-scale_2018}.  Initially, DNNs required supervision on large labeled datasets \cite{krizhevsky_imagenet_2017, deng_imagenet_2009} and did not transfer broadly to downstream tasks. These issues motivated a shift to self-supervised models, which learn representations by grouping similar or temporally-proximate images  \cite{wu_unsupervised_2018, zhuang_local_2019, chen_simple_2020, grill_bootstrap_2020, tong_videomae_2022, caron_emerging_2021, bardes_revisiting_2024, assran_v-jepa_2025}.  Such self-supervised models also turn out to provide a description of neural and cognitive patterns in the primate visual system, rivaling or exceeding the accuracy of the earlier supervised systems \cite{zhuang_unsupervised_2021, konkle_self-supervised_2022}. 

However, from a developmental point of view, modern self-supervised learning methods are a ``glass half full''.  When trained on the visual data diet of real infants and children, they achieve substantial improvements compared to earlier methods such as predictive coding \cite{zhuang_unsupervised_2021, orhan_self-supervised_2020, lotter_neural_2020}, but 
still are far from matching human abilities and perform much worse than similar methods trained on curated, highly non-natural image databases such as ImageNet~\cite{orhan_self-supervised_2020, zhuang_unsupervised_2021, sheybani_curriculum_2023, orhan_self-supervised_2024, long_babyview_2024}. It has remained very difficult for even the best visual learning algorithms from the AI literature to efficiently extract powerful representations from natural data, perhaps due to camera motion/blur, occlusions, and the relatively low diversity of environments children encounter \cite{clerkin_real-world_2017, clerkin_real-world_2022, tan_assessing_2025, sepuri_characterizing_2025, yang_quantifying_2025}. This \emph{ecological data learning gap} is also strikingly present for natural language processing, where large language models (LLMs) need large-scale, highly-curated data to achieve linguistic competency \cite{warstadt_findings_2023, warstadt_call_2023, frank_bridging_2023}. 

Equally important, while current self-supervised visual models may learn useful representations, they cannot perform tasks directly in the flexible zero-shot manner of humans. Instead, for each downstream task, a separate readout must be trained using labeled data, making the overall pipeline for any given task (e.g., image segmentation) ecologically implausible. By contrast, in the language domain, modern LLMs can perform diverse tasks flexibly and zero-shot -- although this behavior only emerges with extremely large amounts of data. The dual challenges of data efficiency and flexibility reflect our incomplete understanding of what inductive structure to build into our models to match human cognition. 

Here, we report substantial progress on the algorithmic foundations of data efficiency and flexibility, by building the Zero-shot World Model (ZWM), a self-supervised neural network that flexibly performs a broad suite of visual-cognitive tasks zero-shot, i.e., without task-specific training/examples (Figure \ref{fig:main_framework}B). ZWM is based on three key principles (Figure \ref{fig:main_framework}A). First, ZWM's underlying learned component is a \emph{sparse temporally-factored} predictor model, a neural network that learns to make predictions given sparse and variable amounts of information. 
Second, after training, the model can be manipulated to yield a universal zero-shot prompting interface by comparing predictions under ground-truth inputs to predictions under minimally modified inputs -- a form of approximate causal inference. Third, ZWM composes simple prompts into more complex queries -- for example, simulating hypothetical motions of objects and then computing the resulting optical flow to segment those objects -- thus, building a computational graph of visual representations that progressively extracts and integrates increasingly complex structure.  Taken together, these components constitute a kind of data-driven world model that is able to forecast the effect of proxy actions on the visual scene.

To test the developmental hypothesis of the ZWM framework and probe its data efficiency under naturalistic conditions, we leverage the BabyView dataset \cite{long_babyview_2024}, a set of egocentric video recordings from young children aged $\sim$5 months to 5 years. Trained solely on 868 hours of BabyView videos ($\sim$3 months of waking experience), the BabyZWM model performs competitively with supervised, state-of-the-art models on challenging real-world datasets across a wide variety of visual cognitive tasks despite receiving no task-specific supervision or labeled probes. Beyond strong behavioral performance, \changes{BabyZWM exhibits a developmental trajectory with progressive emergence of visual-cognitive capacities, and internal representations that align with biological neural responses.}
Taken together, our results suggest one potential answer to the long-standing question of how general-purpose, zero-shot visual cognition can emerge from limited experience -- and, simultaneously, a new approach to building data-efficient AI systems that learn flexibly from limited, uncurated data.

\section*{The ZWM framework}
The Zero-shot World Model (ZWM) concept operationalizes our approach to data-efficient, zero-shot visual world modeling in three design principles (Figure \ref{fig:main_framework}A): temporally-factored prediction to flexibly separate appearance from dynamics; zero-shot extraction of visual-cognitive structures from the predictor through approximate causal inference; and composing extractors together to achieve increasingly complex inference abilities. Taken together, the three components of ZWM form a kind of data-driven ``world model'' -- so-called because the system can be used to forecast the effect of proxy actions (the probes) on a scene. 

\paragraph*{Sparse temporally-factored prediction.}
The core learned component of the ZWM is a \textit{sparse temporally-factored} masked multi-frame visual predictor, which we will denote by $\Psi$. Though the concept can be applied in longer-range many-frame videos (or even non-visual data domains), it is easiest to understand in the two-frame setting we use here (Figure \ref{fig:main_framework}A). 
Given two RGB video frames $f_1$ and $f_2$ separated by a short time gap, let $\Psi_{\Theta}$ be a parameterized function that seeks to predict $f_2$ from $f_1$ together with a small fraction of pixel patches from $f_2$.  The inputs to $\Psi_{\Theta}$ are thus of the form $(f_1, f_2^{\text{masked}})$, where $f_2^{\text{masked}}$ is a subset of the patches of $f_2$ after a mask has been applied.  $\Psi_{\Theta}$ then outputs an estimate of the whole of $f_2$, denoted $\widehat{f}_2$. During training on ground truth frame pairs, the parameters $\Theta^*$ are optimized to minimize the average L2-loss of the prediction across the training dataset $\mathcal{D}$:
\begin{equation}
\Psi_{\Theta^*}: \left(f_1, f_2^{\text{masked}}\right) \longmapsto \widehat{f}_2; \quad \quad \Theta^* := \arg \min_{\Theta} \left \langle \lVert f_2 - \widehat{f}_2 \rVert^{2}\right \rangle_{(f_1, f_2) \in \mathcal{D}}.
\end{equation}
Two critical aspects of the training of $\Psi$ are that (i) the masks applied are very sparse -- e.g. no more than 10\% of the patches of $f_2$ are revealed; and (ii) 
training is performed with randomly-chosen masks, requiring no situation-specific knowledge of the semantic contents of the frames. $\Psi$ is a type of \emph{masked autoencoder}~\cite{bear_unifying_2023, he_masked_2021}, in which the training mask is temporally biased to reveal all of one frame ($f_1$) and very little of the other ($f_2$).  The constraints on this prediction problem are very generic -- merely that masked training is performed, and that the masks are temporally biased.  However, it turns out that this apparently weak generic constraint forces $\Psi$ to learn a very structured representation of the visual scene. Specifically, to successfully reconstruct $f_2$, $\Psi$ implicitly must:
infer object \emph{appearance} from the dense patches of $f_1$, as the small number of patches revealed from $f_2$ are insufficient to do so, while inferring object and camera \emph{motion transformations} from the sparse revealed patches in $f_2^{\text{masked}}$.
$\Psi$ thus implicitly \emph{factorizes} appearance and motion, compressing low-dimensional motion data into a compact, but naturally interpretable, set of patches. 

\paragraph*{Zero-shot extraction via approximate causal inference.}
A key insight of ZWM is that the highly compressed but interpretable motion patches that the training process creates can be manipulated with ``zero-shot prompts'' to extract key visual quantities by making the trained predictor's implicit knowledge explicitly available (Figure \ref{fig:main_framework}A).  
The core mechanism of this process is to: (i) formulate a \emph{minimal perturbation} of a ground-truth input; and (ii) \emph{compare} the predictor $\Psi$'s output in both the original ground-truth case and the minimally perturbed case, and (iii) \emph{aggregate} the difference to hone in on the quantity of interest. 
Formally, this process can be represented as: 
\begin{equation}
x_{\delta} := \textbf{perturb}(x); \quad \quad \delta \Psi := \textbf{compare}(\Psi(x), \Psi(x_{\delta})); \quad \quad \text{output} := \textbf{aggregate}(\delta \Psi).
\end{equation} 
For example, to segment an object, the \textbf{perturb} function can simply induce hypothetical motion by translating one small patch on the object, causing the predictor $\Psi$ to propagate hypothetical motion to the rest of the object, but not other components in the scene; the \textbf{compare} function computes the optical flow between the perturbed and unperturbed cases; and \textbf{aggregate} thresholds the flow to determine which pixels belong to the object.
We show that a wide variety of visual concepts can be extracted in a zero-shot manner from $\Psi$ by choosing different but extremely simple \textbf{perturb}, \textbf{compare}, and \textbf{aggregate} functions.  


ZWM's approach to zero-shot extraction is a form of \textit{approximate causal inference}.  As discussed in the causality literature~\cite{pearl_2009_causality, gerstenberg_counterfactual_2024}, the process of causal inference asks how an outcome of a dynamic process changes when a minimal change is made to its antecedents.
Analogously, the $\Psi$ function acts as a learned structural equation for the world's dynamics, whose temporally-factored nature permits the construction of minimal perturbations that expose some aspect of the causal structure of the world.
For example, ZWM's object segmentation procedure uses a motion perturbation to expose the underlying causal structure of the world --- groups of pixels move together due to the latent \textit{cause} of belonging to the same physical object.

\paragraph*{Compositional prompting.}
ZWM composes simple prompts to construct more complex queries, progressively extracting and integrating increasingly abstract visual structures, e.g., motion and objects, rather than RGB pixels (Figure \ref{fig:main_framework}A). ZWM (i) estimates optical flow from RGB; (ii) computes optical flow on binocular views for relative depth; (iii) simulates hypothetical motions and computes optical flow to segment objects; and (iv) uses flow and segments for intuitive physics. Consequently, this composition builds a computational graph of visual intermediates.

\paragraph*{Model implementation.} We implement the predictor $\Psi$ as a neural network, and perform learning via stochastic gradient descent on its parameters. The base network architecture is a Vision Transformer (ViT) backbone~\cite{dosovitskiy_image_2021}, with versions at two sizes (170 million and 1 billion parameters). We compare models trained on a variety of datasets, as described below. In each case, training datapoints consist of RGB frame pairs taken from a real-world video distribution, with an inter-frame gap sampled uniformly between 150ms and 450ms. The images are input into the model as square 256x256 pixel arrays, and then patchified into 8x8-pixel patches. During training, masks are chosen randomly on each example, with 10\% of patches in the second frame revealed.  

\section*{Results}


\subsection*{ZWM performs diverse visual-cognitive tasks zero-shot}

How well can ZWM flexibly perform a broad suite of visual-cognitive tasks zero-shot? We evaluate a spectrum of visual-cognitive tasks that humans perform, from lower- to higher-level, including optical flow, relative depth estimation, object segmentation, and intuitive physical reasoning. To test robustness across visual diets, we train ZWM on \textbf{BabyView} (N=34, 868 hours, 2025.1 release) \cite{long_babyview_2024} (BabyZWM), \textbf{Kinetics-400} \cite{kay_kinetics_2017} ($\sim$670 hours) (smaller than BabyView but far more diverse Internet videos), and a \textbf{Big Video Dataset} (BVD) \cite{kotar_world_2025} ($\sim$7000 hours; computer vision datasets and Internet videos; an approximate upper bound on what can be achieved with high visual diversity and scale).

We evaluate ZWM against a range of alternative hypotheses, including both representation-based models and task-specific systems. Representation-based systems learn general-purpose visual features from pretraining that are typically transferred to downstream tasks via finetuning or lightweight task-specific heads/objectives. Unlike ZWM, these models are not natively zero-shot, a key limitation for modeling human vision, so we design simple zero-shot probes to evaluate these models. 
\changes{We sweep each baseline's evaluation hyperparameters (e.g., feature layer and thresholds), reporting its best configuration (Supplementary Tables \ref{tab:flow_methods}, \ref{tab:depth_methods}, \ref{tab:seg_methods}).}
As a \textbf{``standard'' supervised static image model,} we evaluate ResNet-50 trained on ImageNet-1k with category-label supervision. For a \textbf{task-generic self-supervised static image model,} we evaluate DINOv3~\cite{simeoni_dinov3_2025} (off-the-shelf and trained on BabyView), which learns strong single-image representations by training the model to produce consistent features across different views of the same image.
As an example of a \textbf{task-generic self-supervised video model}, we evaluate V-JEPA2~\cite{assran_v-jepa_2025} (off-the-shelf and trained on BabyView), a model that learns by predicting masked regions of a video in feature space rather than in raw pixels. 

Poor results across visual-cognitive tasks for these representation-based comparison models do not imply they do not develop useful and powerful visual-cognitive representations, but rather that these models provide no means of accessing these representations zero-shot. 
We therefore also benchmark ZWM against state-of-the-art task-specific baselines, including models trained directly for individual benchmarks (e.g., supervised networks optimized for flow, depth, or segmentation). These baselines can be viewed as concrete instantiations of the alternative hypothesis that human-like competence is achieved via separate, specialized systems rather than a unified world model. Due to the paucity of benchmarks for direct model-to-human comparisons, and because humans would be expected to perform near ceiling on these everyday visual tasks, we instead treat supervised state-of-the-art systems as a strong proxy baseline. Outperforming them therefore provides a strong test of BabyZWM’s zero-shot data efficiency.


We evaluate on established benchmarks designed to be challenging (real-world motion, occlusions, and lighting changes). For fair comparisons, we provide all models with the same inputs. We describe detailed methods for each task in the Supplementary Materials.

\begin{figure}[p]
	\centering
	\includegraphics[width=1.0\textwidth]{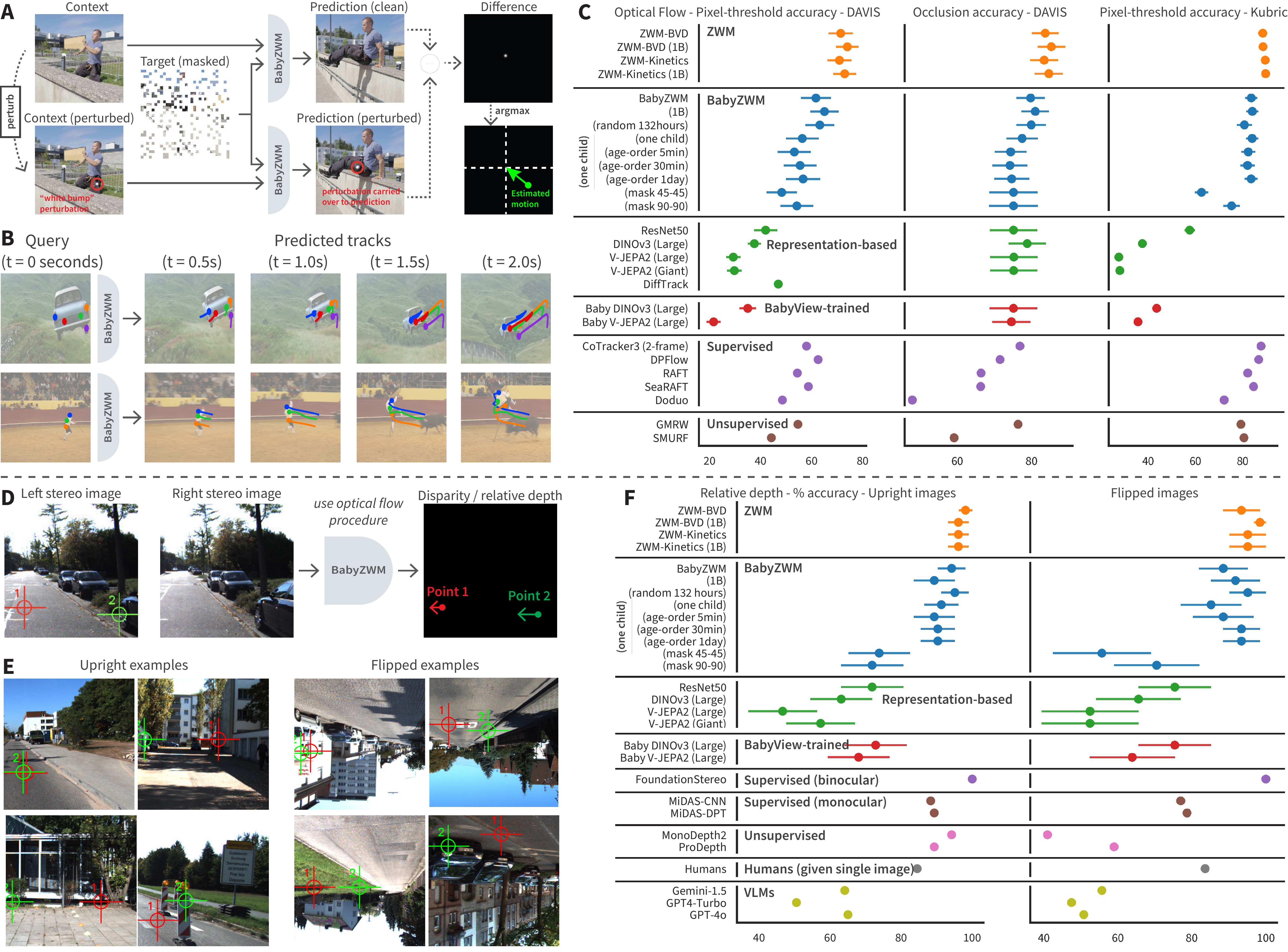}\\
	\includegraphics[width=1.0\textwidth]{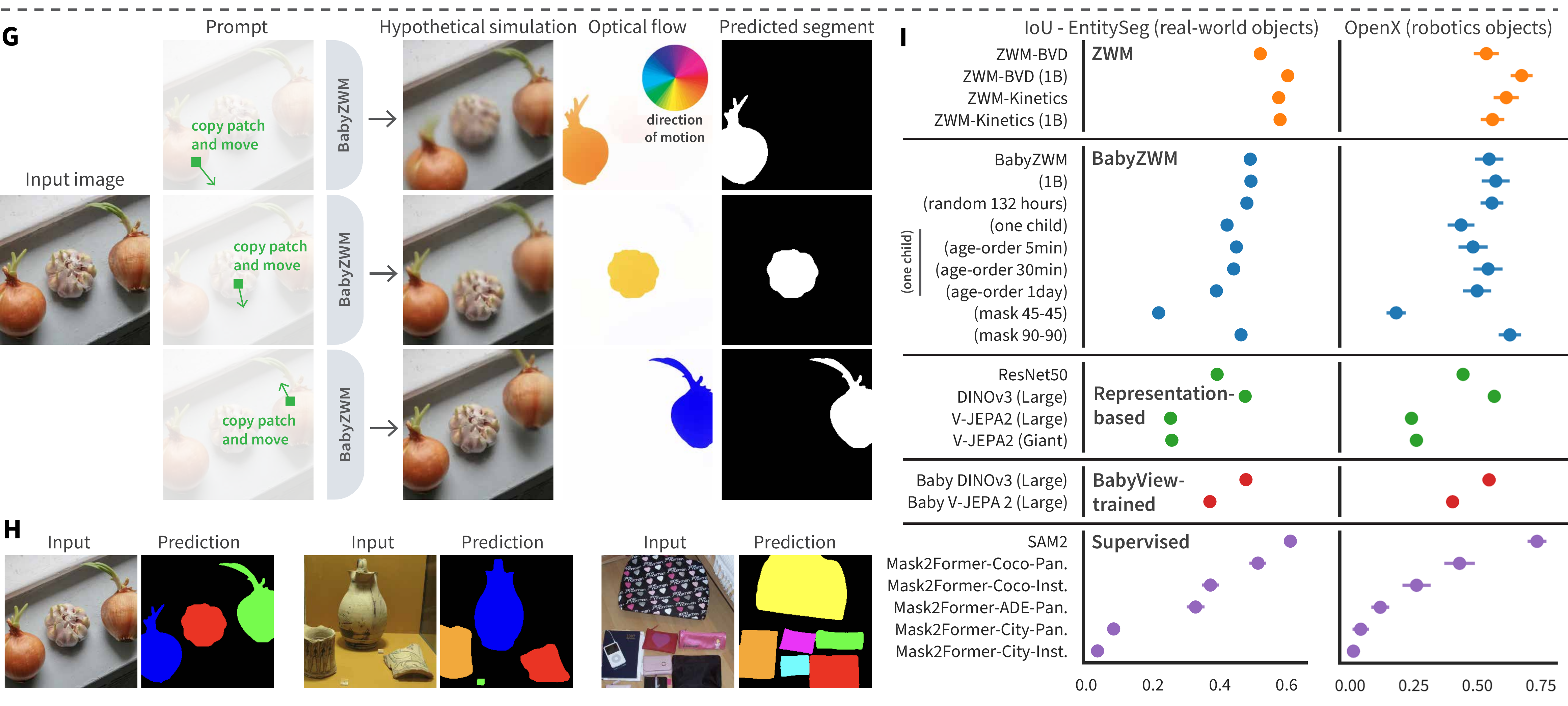}
	\caption{
        \textbf{BabyZWM estimates optical flow, relative depth, and object segments zero-shot.}
        (\textbf{A}) Optical flow method. (\textbf{B}) Flow predictions as tracks. (\textbf{C}) BabyZWM matches state-of-the-art optical flow.
        (\textbf{D}) Relative depth method (built on optical flow). (\textbf{E}) Benchmark examples (upright and flipped). (\textbf{F}) BabyZWM beats supervised monocular depth models.
        (\textbf{G}) Object segmentation and hypothetical motion procedure. (\textbf{H}) Segmentation predictions. (\textbf{I}) BabyZWM matches supervised segmenters, except SAM2. Error bars indicate bootstrap 95\% intervals throughout the paper. \changes{Representation baselines are each read at their best configuration over a layer $\times$ threshold sweep (Supplementary Table \ref{tab:seg_methods}).}
    }
	\label{fig:main_flow_depth}
	\label{fig:main_segmentation}
\end{figure}

\paragraph*{Optical flow.} To rigorously evaluate algorithm performance on optical flow, we use several recent computer vision benchmarks: TAP-Vid-DAVIS \cite{doersch_tap-vid_2023}, consisting of challenging real-world videos with human-annotated ``ground-truth''; and TAP-Vid-Kubric \cite{greff_kubric_2022}, consisting of synthetic, simulator-generated videos where ground-truth flows are known by construction. \changes{To enable controlled comparisons, we pass in two frames (query, target) to all models.} We measure pixel-threshold accuracy (percentage of predictions within a pixel-radius of ground truth) and occlusion/out-of-frame detection accuracy. 
ZWM achieves state-of-the-art results (Figure \ref{fig:main_flow_depth}C); BabyZWM is competitive with label-supervised CoTracker3 \cite{karaev_cotracker_2024}, DPFlow \cite{morimitsu_dpflow_2025}, and SeaRAFT \cite{wang_sea-raft_2024} on TAP-Vid-DAVIS and matches supervised baselines at detecting occlusions. On TAP-Vid-Kubric, BabyZWM is strong but slightly below supervised models (which use synthetic training). BabyZWM outperforms the DINOv3 and V-JEPA2 models.

\paragraph*{Relative depth estimation.} We evaluate depth perception on UniQA-3D \cite{zuo_towards_2024}: point pairs that require judging which is further. 
Depth is extracted zero-shot from ZWM by computing optical flow between stereo images (Figure \ref{fig:main_flow_depth}D).
Both ZWM and BabyZWM exceed 90\% accuracy (Figure \ref{fig:main_flow_depth}F). They surpass large vision–language models (Gemini-1.5, GPT-4-Turbo, GPT-4o) \cite{gemini_gemini_2024, openai_gpt-4_2024, openai_gpt-4o_2024}, are comparable to supervised (MiDaS-CNN \cite{ranftl_towards_2020}) and self-supervised (MonoDepth2 \cite{godard_digging_2019}) monocular estimators, and trail only a supervised binocular model \cite{wen_foundationstereo_2025}.

\paragraph*{Object discovery.} We evaluate object segmentation on SpelkeBench \cite{venkatesh_discovering_2025}, a class-agnostic benchmark defining objects as distinct, bounded entities.
On this benchmark, BabyZWM rivals supervised Mask2Former variants \cite{cheng_masked-attention_2022} trained on large-scale COCO \cite{lin_microsoft_2015}, though it performs slightly below SAM2 \cite{ravi_sam_2024} which leverages large-scale human annotations (Figure \ref{fig:main_segmentation}I). 

\begin{figure}[tbp]
	\centering
	\includegraphics[width=0.9\textwidth]{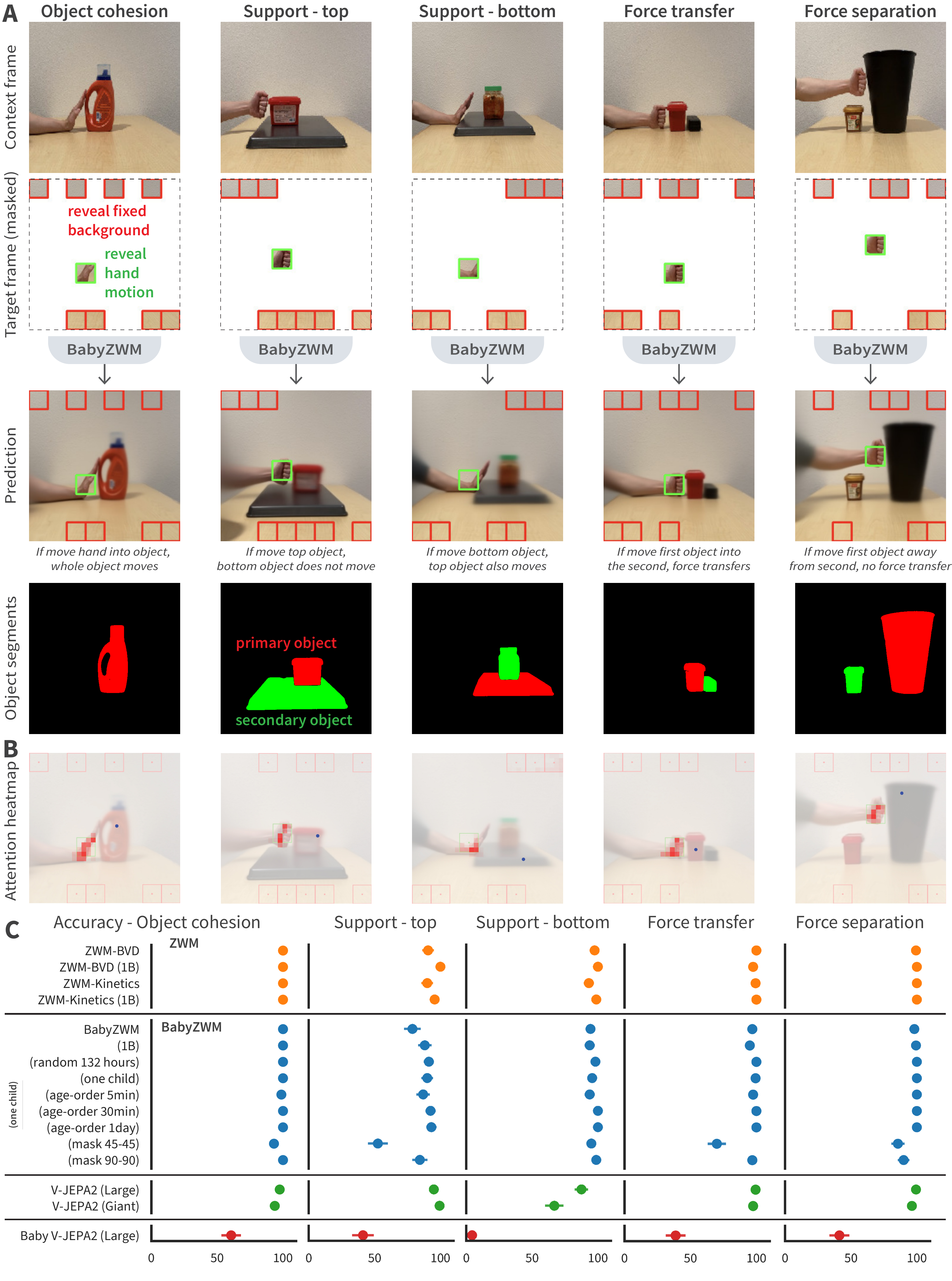}
	\caption{
        \textbf{BabyZWM exhibits object knowledge and intuitive physics.}
        (\textbf{A}) Short-timescale benchmark testing cohesion, support (move top/bottom object), force transfer, and force separation; given a context frame and a few target patches, predict remaining patches. \changes{We reveal the hand's target location, and fixed border patches to signal a static camera.}
        (\textbf{B}) Interpretability methods reveal attention heads that track the hand (causal agent) when predicting the target object (marked with a blue point).
        (\textbf{C}) ZWM, BabyZWM, and V-JEPA2 near 100\% on most categories; Baby V-JEPA2 does not.
    }
	\label{fig:main_object_intuitive_physics}
\end{figure}

\paragraph*{Intuitive physical understanding.} We develop a novel short-timescale physical reasoning benchmark (Figure \ref{fig:main_object_intuitive_physics}A) to evaluate models, featuring interactions between a hand and 1–2 objects that test 5 categories of reasoning: object cohesion, support relations with motion of either the top or bottom object, force transfer, and force separation.
\changes{A prediction is counted as correct if it is closer to the ground-truth than the context frame}, using mean squared error and LPIPS perceptual similarity \cite{zhang_unreasonable_2018} \changes{(Methods)}. 
ZWM, BabyZWM and V-JEPA2 approach 100\% performance on almost all categories, but not Baby V-JEPA2 (Figure \ref{fig:main_object_intuitive_physics}C). We apply model interpretability techniques to BabyZWM, revealing several attention heads that consistently follow the hand (causal agent) when predicting the motion of the object of interest (Figure \ref{fig:main_object_intuitive_physics}B; \changes{Supplementary Figure \ref{fig:supp_attention_heads}}). \changes{Randomly selected qualitative predictions for every category are shown in Supplementary Figure \ref{fig:supp_physics_qualitative}.}
\changes{The benchmark does not test object permanence, which requires multi-frame memory, nor mass sensitivity, since a hand-driven push moves objects regardless of mass. A preliminary qualitative test of the latter is reported in the Supplement (Figure~\ref{fig:supp_mass_sensitivity}).}

\subsection*{ZWM achieves data efficiency and continual learning}

A correct theory of visual-cognitive learning must be able to achieve effective learning on the real datastreams that a human experiences. 

BabyZWM
retains most of its performance compared to the same architecture trained on much more diverse datasets, such as Kinetics-400 and BVD (Figures \ref{fig:main_flow_depth}, \ref{fig:main_object_intuitive_physics}), emphasizing the data-efficiency of the ZWM architecture.


To perform an even more stringent test, we next trained ZWM on 
\textbf{Single-Child BabyView}, a subset of BabyView consisting of 132 hours of recordings from a single individual (age 9--30 months). The Single-Child dataset represents a stricter test for model learning, because it requires algorithms to be able to learn generalizable capacities from the highly restricted visual diversity of one child's experience.  (We additionally train on a random 132-hour subset, allowing us to disentangle the contributions of diversity versus total exposure.)
Single-Child BabyZWM performs similarly to BabyZWM across most tasks (Figures \ref{fig:main_flow_depth}, \ref{fig:main_object_intuitive_physics}). 

Additionally, we trained a version of Single-Child BabyZWM on a single pass through a version of the data in which the video clips were ordered by the child's age. This is an important test of developmental robustness and continual/life-long learning, \changes{a setting also studied outside of developmental contexts \cite{carreira_learning_2024}}.
We create a set of curricula by shuffling within various temporal windows (5 minutes, 30 minutes, and 1 day), loosely approximating different degrees of experience consolidation (e.g., within-episode mixing vs. sleep-like reordering). The age-ordered Single-Child BabyZWM models performed similarly to Single-Child BabyZWM across all tasks (Figures \ref{fig:main_flow_depth}, \ref{fig:main_object_intuitive_physics}). 

Finally, ``Standard'' BabyZWM uses asymmetric masking (fully visible $f_1$, 90\% masked $f_2$), explicitly prioritizing the learning of motion dynamics. Because this temporally-factored mask structure contributes a conceptually core component of the ZWM concept, we explore simpler alternatives. We evaluate symmetric masking variants of BabyZWM (mask 45\%-45\% and mask 90\%-90\%), which perform substantially worse (Figures \ref{fig:main_flow_depth}, \ref{fig:main_objective_readout}), showing that emphasizing motion information is useful for data efficiency and zero-shot abstraction.

\subsection*{\changes{Representation and readout are both important}}

\begin{figure}[!t]
    \centering
    \includegraphics[width=1.0\textwidth]{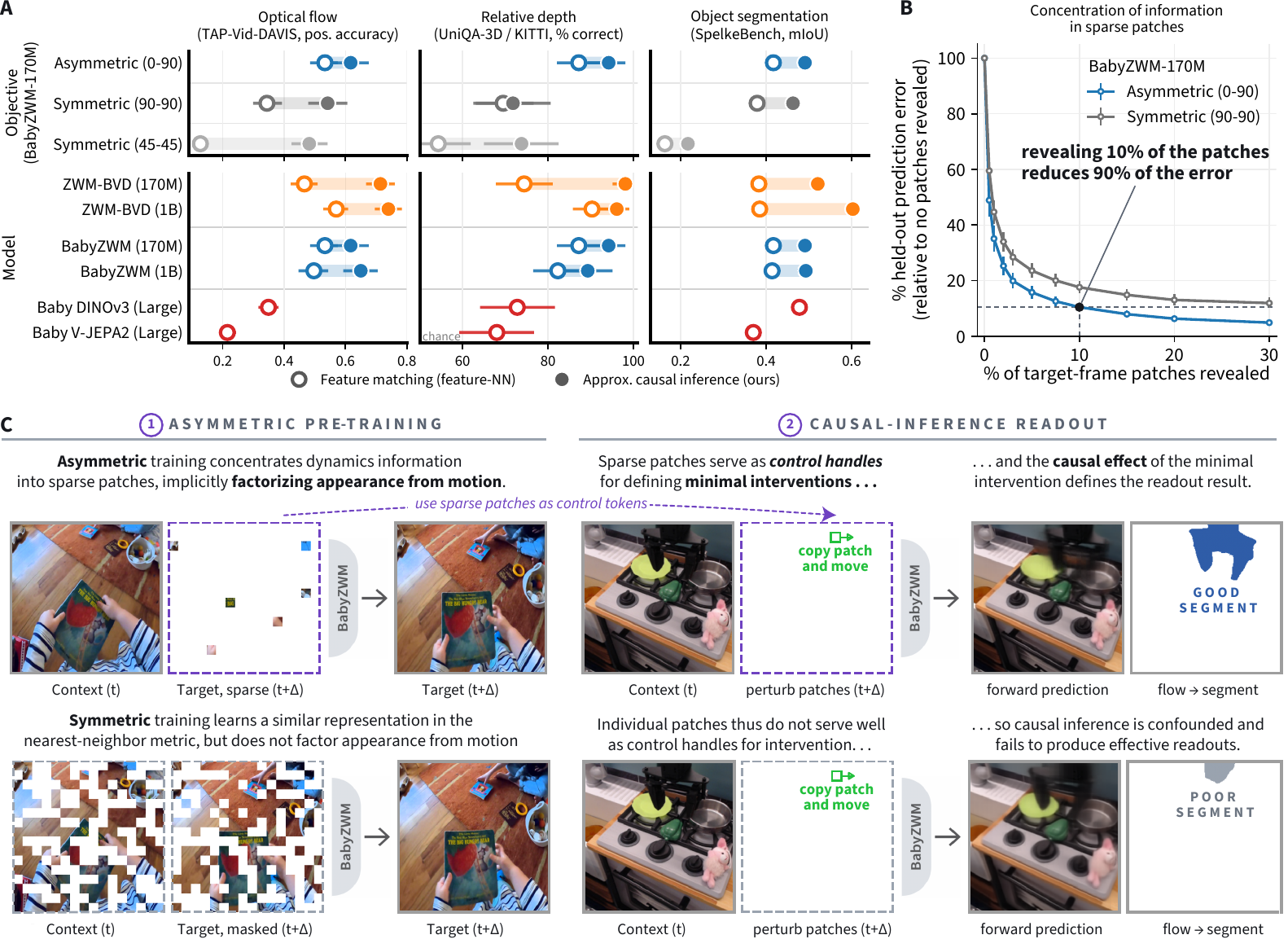}
    \caption{\changes{
        \textbf{Representation and readout are both important to ZWM’s success.}
        (\textbf{A}) Objective $\times$ readout for BabyZWM-170M (top), and readouts across models (bottom). Both representation-learning choices and read-out method matter; degrading either costs performance on every task.
    (\textbf{B}) With the context frame fully visible, we reveal different proportions of target-frame patches and score prediction error on 20\% of patches never revealed -- across 100 Kinetics-400 videos (Supplementary Table \ref{tab:patch_sufficiency}). Revealing 10\% of patches reduces 90\% of the error -- sparse patches carry most of the target frame's predictive information (how camera and objects move). Asymmetric training concentrates this information more than the symmetric ablation.
        (\textbf{C}) (\emph{Top row}) Asymmetric (ZWM) versus symmetric training objective. 
        ZWM's temporal-factoring asymmetric objective concentrates dynamics information into sparse patches, which can then be used as control tokens to create minimal interventions for powering causal-inference readout. (\emph{Bottom row}) Symmetric control training objective with the same amount of information
        does \emph{not} produce usable control handles for minimal interventions, confounding causal inference.  
    }}
    \label{fig:main_objective_readout}
\end{figure}

\changes{
A natural question is whether ZWM's zero-shot performance is due to a better learned representation or a better readout mechanism. To answer this question, we crossed the two training objectives (temporally-factored asymmetric masking vs. symmetric masking) with two readout mechanisms (our approximate causal inference vs. the label-free feature nearest-neighbor probe used for the representation baselines). This yields four settings, evaluated on optical flow, relative depth, and object segmentation (Figure \ref{fig:main_objective_readout}; Supplementary Table \ref{tab:mask_ablation}). Notably, asymmetric masked autoencoding read out by feature-matching is the setting closest to prior work, in particular SiamMAE \cite{gupta_siamese_2023}.

Neither factor alone accounts for ZWM's performance: the full combination is the best of the four settings on every task, and degrading either factor costs performance on every task.  Read with the feature nearest-neighbor probe, the BabyZWM representation performs at roughly the level of state-of-the-art self-supervised baselines -- above DINOv3 and V-JEPA2 on flow and depth, but below DINOv3 on segmentation.  Yet, when read with our approximate causal-inference mechanism, BabyZWM alone becomes competitive with state-of-the-art \emph{task-specialized supervised} solutions (Figure \ref{fig:main_flow_depth}). 

We investigated the specific mechanism underlying this result (Figure \ref{fig:main_objective_readout} B-C).  We compared a symmetric-trained model given the same total amount of information as the asymmetric ZWM objective, but in which the information was distributed differently (that is, symmetrically) during training. This is closer to the ``usual'' VideoMAE masking objective~\cite{tong_videomae_2022}.  The representation thus trained is similar in nearest-neighbor geometry, but has drastically worse causal-readout performance.  This is because, since the symmetric objective does not concentrate dynamics information in sparse tokens, minimal interventions are less effective, causal inference is heavily confounded, and the readout quantities are poor.  

The temporally-factored objective therefore builds the correspondence structure that the causal-inference readout exploits, and the readout is what converts that structure into task performance.   Under this view the readout is part of the model in a deep way -- a point that matters especially for a model of child development, where no task labels are available to train a readout.
}

\begin{figure}[tbp]
	\centering
	\includegraphics[width=1.0\textwidth]{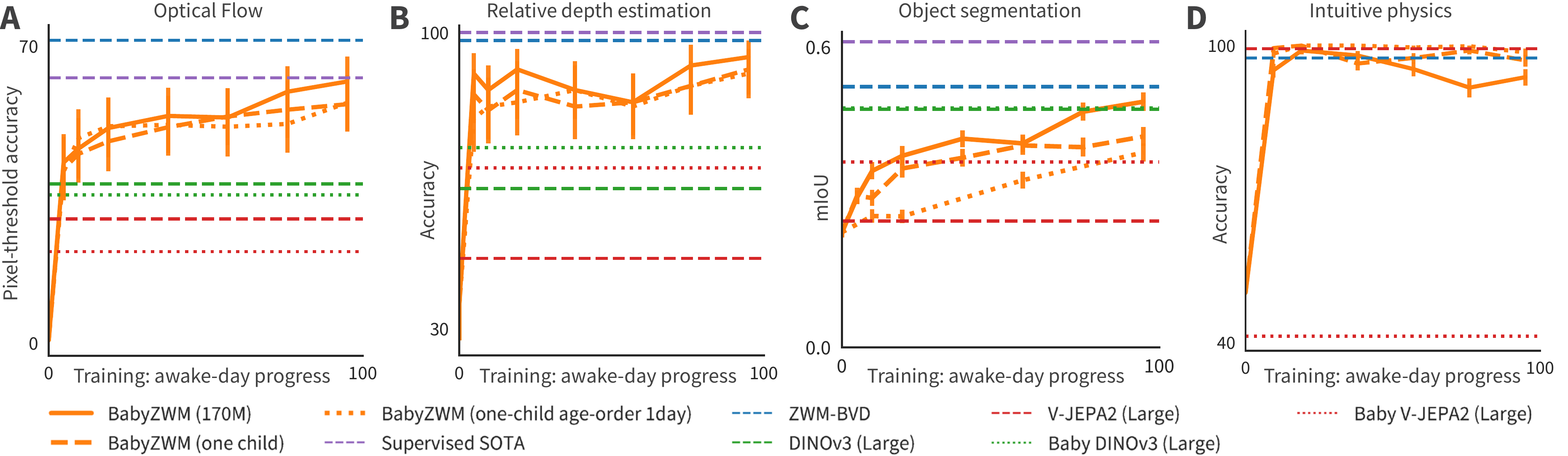}
	\caption{
        \textbf{BabyZWM develops zero-shot capacities across training checkpoints}. We plot the developmental trajectories of BabyZWM, Single-Child BabyZWM, and Single-Child BabyZWM (age-order, shuffle within each day) to observe how fast different visual-cognitive capacities emerge. We evaluate these models across a full training run, which corresponds to roughly 95 days of waking experience assuming $\sim$10 awake hours/day \cite{iglowstein_sleep_2003}. We also compare these to ZWM trained on BVD, supervised state-of-the-art baselines, and other alternative hypotheses.
        We plot these developmental trajectories for
        (\textbf{A}) optical flow,
        (\textbf{B}) relative depth estimation,
        (\textbf{C}) object segmentation, and
        (\textbf{D}) intuitive physical reasoning.
    }
	\label{fig:main_dev_trajectories}
\end{figure}

\subsection*{\changes{Progressive, staged emergence of visual-cognitive capacities}}

Having evaluated the BabyZWM models, we next ask what we can learn from looking at their developmental trajectories and when different visual-cognitive capacities emerge.

\changes{BabyZWM's optical flow and object segmentation capabilities increase across the entire training period, whereas depth and intuitive physical capacities rise steeply and peak early (Figure \ref{fig:main_dev_trajectories}).}
\changes{We measure each capacity's emergence over its \emph{learnable gain} -- the rise from a floor (chance for the two-alternative depth and physics benchmarks, the value at initialization otherwise) to its maximum across checkpoints. We report the percentage of training needed to reach 80\% of that gain (also 50\%; Supplementary Table \ref{tab:emergence_timing}). Relative depth emerges earliest (4.6\% of training, to 80\% gain), with intuitive physics next (8.7\%), optical flow later (17.4\%), and object segmentation latest (69.8\%).}

These comparisons should be interpreted cautiously: they partly reflect benchmark-specific design choices -- especially differences in task difficulty, metrics, and ceiling effects -- rather than a clean ordering of underlying capability development. \changes{What the model does supply is quantitative, testable predictions of the rate at which each capacity emerges. These emergence dynamics can be compared against the order and timing of capacity development in children -- object tracking \cite{trick_multiple-object_2005, blankenship_development_2020}, early stereopsis \cite{held_stereoacuity_1980, fox_stereopsis_1980, birch_stereoacuity_1982, norcia_late_2025}, object segmentation and intuitive physics \cite{johnson_how_2010, baillargeon_object_2012, baillargeon_development_1992, hespos_reasoning_2001}. However, adjudicating that comparison requires systematic, matched developmental benchmarks for humans and models, which is an important future direction.}

\begin{figure}[tbp]
	\centering
	\includegraphics[width=1.0\textwidth]{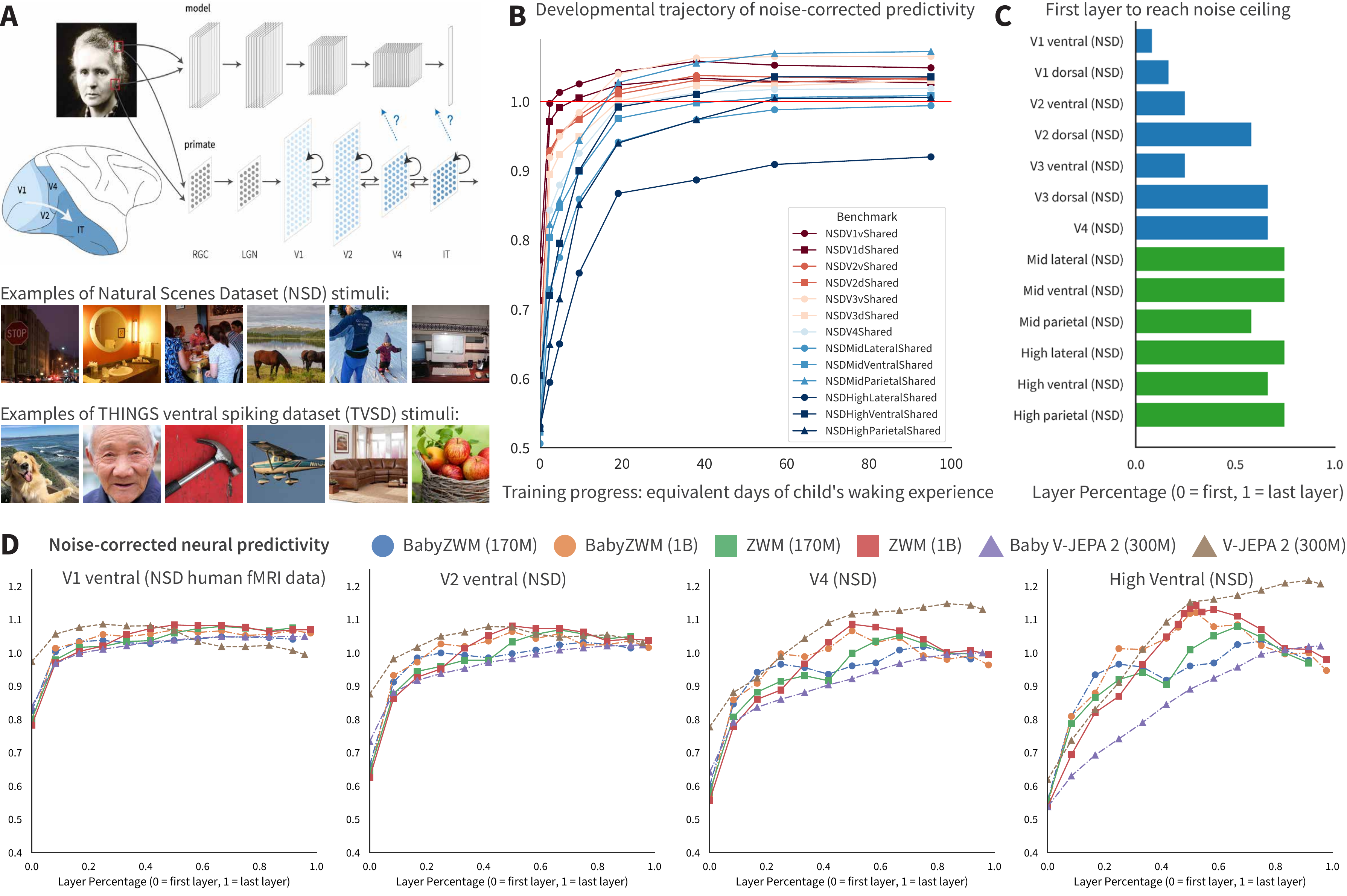}
	\caption{
        \textbf{BabyZWM successfully develops internal representations that align with neural responses from human fMRI and macaque electrophysiology datasets.}
		(\textbf{A}) Neural predictivity schematic \cite{schrimpf_brain-score_2018}, with example images from NSD and TVSD.
        (\textbf{B}) Developmental trajectory: BabyZWM's neural predictivity for early visual areas increases quickly in training, while it takes longer for the later areas, exhibiting an ``early-first'' developmental trajectory. We observe this both in the steeper slope for neural predictivity of V1 than higher regions, as well as neural predictivity reaching V1's noise ceiling at an earlier checkpoint.
        (\textbf{C}) For various visual areas in the brain, we plot the first layer of BabyZWM that reaches the noise ceiling. For earlier cortical regions, earlier model layers reach the noise ceiling, whereas later cortical regions align with deeper layers. This exhibits neuroanatomical consistency with several accounts of hierarchical visual organization.
        (\textbf{D}) Detailed plots for noise-corrected neural predictivity for the ventral stream for NSD human fMRI.
    }
	\label{fig:main_brain}
\end{figure}

\subsection*{ZWM representations align with neural responses}

\changes{Having characterized ZWM's behavioral trajectories, we next ask whether ZWM also develops brain-like internal representations.} 
The human visual system is organized hierarchically, transforming retinal inputs into increasingly complex representations \cite{felleman_distributed_1991, dicarlo_how_2012, grill-spector_functional_2014} that
develop over childhood \cite{gogtay_dynamic_2004, golarai_differential_2007, grill-spector_developmental_2008, braddick_development_2011, kiorpes_visual_2015}. 
We evaluate the similarity of our models' internal representations (across various training checkpoints) with brain responses by computing neural predictivity \cite{yamins_performance-optimized_2014, guclu_deep_2015, yamins_using_2016}: fit a cross-validated linear probe from model features to neural responses, then report noise-corrected correlations (Figure \ref{fig:main_brain}A). We evaluate two complementary benchmarks: the Natural Scenes Dataset (NSD) \cite{allen_massive_2022} for human fMRI and the THINGS Ventral Stream Spiking Dataset (TVSD) \cite{papale_extensive_2025} for macaque electrophysiology. fMRI captures large-scale representational geometry; electrophysiology reveals fine-grained single-neuron tuning/timing.

Across NSD and TVSD, the BabyZWM model exhibits neural alignment consistent with hierarchical visual development. Neural predictivity for early visual cortex approaches its noise ceiling at relatively early training checkpoints, whereas higher regions improve more gradually, an ``early-first'' developmental trajectory (Figure \ref{fig:main_brain}B). Layer–area correspondence is hierarchically aligned: for earlier cortical regions, earlier model layers reach the noise ceiling, whereas later cortical regions align with deeper layers (Figure \ref{fig:main_brain}C). This pattern is consistent with several accounts of hierarchical visual organization \cite{felleman_distributed_1991, goodale_separate_1992, dicarlo_how_2012, grill-spector_functional_2014} and prior modeling findings \cite{yamins_performance-optimized_2014, guclu_deep_2015}, supporting an explicit ``mechanistic mapping'' between model layers and cortical regions \cite{kaplan_explanatory_2011, cao_explanatory_2021, frank_cognitive_2025}. A single, self-supervised world model thus captures representational structure shared across species and measurement scales, and BabyZWM \changes{recapitulates brain-like signatures} of both developmental dynamics and hierarchical organization.

\changes{The ZWM objective is also robust to a developmental data diet. We quantified the change in peak neural predictivity when an architecture is trained on BabyView instead of its standard training set (Figure \ref{fig:main_brain}D; Supplementary Table \ref{tab:baby_data_penalty}). Averaged over higher-order visual areas (NSD V4, mid- and high-ventral; macaque V4 and IT), peak predictivity drops by 10.4\% for V-JEPA2 but only 2.1\% (170M) and 1.5\% (1B) for ZWM -- a five-fold smaller ``baby-data'' penalty. Furthermore, the V-JEPA2 penalty grows monotonically along the hierarchy ($-$3.4\% V1, $-$16.2\% high ventral) while ZWM's stays small and roughly flat across areas.}

\changes{Expanded layer-by-region profiles for both benchmarks, baseline comparisons for the layer-area correspondence and the developmental trajectory, and per-area converged predictivity for every model, are given in the Supplement (Supplementary Figures \ref{fig:supp_neural_predictivity}, \ref{fig:supp_layer_area_allmodels}, \ref{fig:supp_dev_trajectory_baselines}; Supplementary Tables \ref{tab:baby_data_penalty}, \ref{tab:converged_predictivity}).}
\changes{An exciting future direction is to extract features while ZWM runs its causal-inference readouts, rather than from static images, which may yield stronger/distinct neural predictions --- though this likely needs neural datasets with dynamic stimuli that engage these inferences.}


\section*{Discussion}

Modern visual learning algorithms are highly data inefficient when compared to humans, experiencing substantial performance gaps when trained on the real datastreams experienced by human children~\cite{orhan_self-supervised_2020, zhuang_unsupervised_2021, sheybani_curriculum_2023, orhan_self-supervised_2024, long_babyview_2024}.  Here, we describe a novel approach to visual learning, Zero-shot World Modeling, that is able to bridge this gap. 
To acquire diverse visual-cognitive capacities without labels, ZWM represents a shift from the dominant paradigm of representation learning with task-specific readouts to unified, zero-shot world models. In representation learning, each downstream task needs its own labeled readout \cite{wu_unsupervised_2018, zhuang_local_2019, chen_simple_2020, grill_bootstrap_2020, tong_videomae_2022, caron_emerging_2021, bardes_revisiting_2024, assran_v-jepa_2025}, limiting the range of feasible tasks and encouraging overfitting to sparse labels. In contrast, ZWM achieves zero-shot, out-of-distribution generalization to challenging real-world scenes, synthetic simulations, and flipped images.  Moreover, ZWM gains competence even when trained with limited data from one individual child, presented in an online single-epoch fashion.

Beyond its implications for cognitive science, ZWM's zero-shot capability addresses a pressing challenge in AI. Current self-supervised visual models, despite learning rich representations, remain locked behind task-specific labeled readouts -- an expensive and brittle dependency that limits practical deployment. ZWM eliminates this bottleneck: a single learned predictor yields optical flow, depth, segmentation, and physical reasoning zero-shot, through a universal interface. This mirrors the paradigm shift in NLP when LLMs replaced task-specific fine-tuned models -- but ZWM achieves this in vision with orders of magnitude less data. That this is achievable from just hundreds of hours of a single child's naturalistic, uncurated video -- rather than millions of hours of curated internet data -- suggests that the right inductive structure can dramatically reduce the data requirements for broad visual competence. This has direct relevance for domains such as robotics, medical imaging, and embodied AI, where large-scale labeled data is unavailable.

ZWM is a natural hybrid between two polar concepts of the role of intermediate structure in cognition and learning. The first is a ``pure learning'' alternative, embodied by Richard Sutton's Bitter Lesson~\cite{sutton_bitter_lesson_2019} -- that complex hand-built inductive biases are unnecessary in formulating effective learning machines.  
The second is the idea, emanating from computational cognitive science, that human learning is best understood as embodying strong priors about the world, 
citing the sophistication and early emergence of infants’ object and physical knowledge \cite{spelke_origins_1992, spelke_core_2000, carey_origin_2009, spelke_what_2022} and poverty-of-stimulus claims that children’s input is too noisy to support learning \cite{chomsky_rules_1980, roberts_argument_2016}. 
The ZWM principles draw on both of these ideas, illustrating how explicit structure can be created within a minimally-biased learned network.  
\changes{Scale suffices when data is abundant, but when it is finite, as in developmental experience, built-in structure is the right design. Such structure can itself be found by search and learning (possibly instantiated by evolution), making it a product, rather than a violation, of the Bitter Lesson.}

The fact that ZWM can implement this hybrid, and the observation that doing so leads to substantial gains in learning efficiency, has implications for the long-standing debate between developmental nativism and empiricism. 
Specifically, ZWM instantiates \changes{a form of ``architectural nativism'' \cite{elman_rethinking_1996} that includes readout programs: a small, fixed scaffold is} innate -- architecture, learning algorithm, and task-specific readout programs -- while the representational content and network parameters are learned from experience. \changes{The readouts are ``zero-shot'': they specify \emph{how} to perform a task without supplying labels/examples. Any readout---including the feature-matching probes applied to the representation baselines---encodes an instruction of this kind.}
Importantly, our results provide proof-of-concept validation that this mechanism supports acquisition of visual-cognitive capacities and object- and physics-like representations from naturalistic visual experience, challenging strong nativist accounts that posit extensive innate biases for representational content and concepts.

Under this interpretation, zero-shot readouts may correspond to evolutionarily-specified, hard-wired neural circuits that map learned dynamics to visual-cognitive percepts. \changes{This scaffold is compact enough to pass through the genomic bottleneck (compact programs, not stored knowledge) \cite{zador_critique_2019, shuvaev_encoding_2024}.} Future work can explore if they might alternatively be learned during development as flexible adapters, or constructed online as query-like cognitive inference routines over the learned predictor.

ZWM achieves zero-shot visual cognition by being a world model, which forecasts the consequences of actions.  This concept has a long tradition within model-based reinforcement learning~\cite{sutton_integrated_1990, Deisenroth2011PILCOAM} and model-predictive control~\cite{garcia_model_1989, ha_world_2018, Hafner2019DreamTC, Hafner2018LearningLD, Hafner2022DeepHP, Kaiser2019ModelBasedRL, Schrittwieser2019MasteringAG, Wu2022DayDreamerWM, Kotar_2023_ICCV}.  
It might at first seem odd that we discuss ZWM as a world model---after all, the inputs to $\Psi$ are just data, so where are the actions whose consequences are to be forecasted? ZWM is a ``data-driven world model'', in which expensive-to-obtain true action data is proxied by cheap data (e.g. pixel-patch) operations that  approximate simple actions -- the ``tracers'' and ``motions'' used to create hypotheticals for computing flow, object segments, etc.  
ZWM formally treats data patches in the same way that true action data would be, and reaps the reward of doing so, because training on raw data enables the underlying model to learn enough about the way the world works that it can competently perform hypotheticals. 
Future work could seek to learn an interactive \emph{policy} for choosing such ``actions'', setting up comparisons to observed child hand and head motions captured in the BabyView dataset.

Our present work has a number of important limitations. First, by focusing on physically-grounded quantities that are learned by very young infants, ZWM leaves unaddressed how semantic concepts -- e.g. named linguistic categories of objects, relationships, activities -- arise developmentally. We hope that future work will integrate the world model learned by ZWM with the rich linguistic/auditory data experienced by children.
Second, a core empirical limitation of the present work is the paucity of detailed developmental behavioral and neural comparisons. 
Such datasets are very challenging to produce and will require concerted collaborative efforts.
Finally, as a deterministic regression model, ZWM's $\Psi$ predictor is subject to \emph{mode collapse}, leading to blurry predictions in situations in which there is underlying uncertainty about how the future will resolve. This design limits our ability to study longer-horizon prediction and control; extending to multi-frame training, richer temporal memory, and long-horizon tasks \changes{(e.g., object permanence)} is an important next step \cite{kotar_world_2025}.


An intriguing line for future work is \emph{integrating} ZWM's zero-shot task extractions into its underlying predictor $\Psi$, so $\Psi$ can be conditioned on, and make predictions of, these intermediate quantities. Recent work in world modeling has suggested a possible mechanism~\cite{kotar_world_2025, lee_3d_2025, kim_taming_2025}, creating a bootstrapping cycle in which every additional intermediate could contribute a learnable target for enriching the predictor, in turn enabling increasingly efficient learning and the identification of more sophisticated intermediates. This might pave the way for even more flexible, data-efficient learning.



\section*{Acknowledgments}
We are very grateful to Cameron Ellis, Hyowon Gweon, James (Jay) McClelland, Cliona O'Doherty, and Alison Gopnik for helpful feedback on our manuscript.

\paragraph*{Funding:}
This work was supported by the following awards. D.L.K.Y.: Simons Foundation grant 543061, National Science Foundation CAREER grant 1844724, National Science Foundation Grant NCS-FR 2123963, Office of Naval Research grant S5122, ONR MURI 00010802, ONR MURI S5847, and ONR MURI 1141386 - 493027.
We also thank the Stanford HAI, Stanford Data Sciences, Stanford Marlowe team, and the
Google TPU Research Cloud team for computing support.

\paragraph*{Author contributions:}
K.L.A., M.C.F., and D.L.K.Y. designed research, analyzed data, and wrote the paper; K.L.A., K.K., and W.L. implemented and trained models; S.K. implemented optical flow algorithms and analyses; K.J. implemented neural predictivity algorithms and analyses; L.N.C. and R.V. implemented object segmentation algorithms and analyses.

\paragraph*{Competing interests:}
There are no competing interests to declare.

\paragraph*{Data and materials availability:}
\changes{Code for reproducible model training and evaluation is available at \url{https://github.com/awwkl/ZWM}. Trained model checkpoints and our intuitive-physics benchmark are released at \url{https://huggingface.co/awwkl}. BabyView is available to researchers through Databrary at \url{https://www.databrary.org/volume/1882}.}

\clearpage


\bibliographystyle{unsrtnat_etal}
\bibliography{baby_model,baby_model_imported}

\newpage

\renewcommand{\thefigure}{S\arabic{figure}}
\renewcommand{\thetable}{S\arabic{table}}
\renewcommand{\theequation}{S\arabic{equation}}
\renewcommand{\thepage}{S\arabic{page}}
\setcounter{figure}{0}
\setcounter{table}{0}
\setcounter{equation}{0}
\setcounter{page}{1} 


\section*{Supplementary Materials for\\ \scititle}

\begin{center}
Khai~Loong~Aw$^{\ast}$,
Klemen~Kotar,
Wanhee~Lee,
Seungwoo~Kim,
Khaled~Jedoui,
Rahul~Venkatesh,
Lilian~Naing~Chen,
Michael~C.~Frank,
Daniel~L.K.~Yamins\\
\small$^\ast$Corresponding author. Email: khaiaw@stanford.edu
\end{center}


\newpage


\section*{Methods}

\subsection*{Model architecture}

The ZWM predictor $\Psi$ is implemented as a Vision Transformer (ViT)~\cite{dosovitskiy_image_2021}. Input frames are resized to $256 \times 256$ pixels and divided into non-overlapping $8 \times 8$-pixel patches, yielding $32 \times 32 = 1024$ patch tokens per frame. We evaluate two model sizes (Table~\ref{tab:architecture}):

\begin{itemize}
    \item \textbf{ZWM-170M}: 24 transformer layers, 12 attention heads, embedding dimension 768, totaling $\sim$170 million parameters.
    \item \textbf{ZWM-1B}: 48 transformer layers, 16 attention heads, embedding dimension 1280, totaling $\sim$1 billion parameters.
\end{itemize}

\paragraph{Two-frame input tokenization.}
Given a frame pair $(f_1, f_2)$, the first frame $f_1$ is fully patchified into 1024 tokens, each a flattened $8 \times 8 \times 3 = 192$-dimensional vector. The second frame $f_2$ is masked: only 10\% of its patches (approximately 102 tokens) are revealed, with the remaining patches replaced by a shared learnable mask token. Both sets of tokens receive positional embeddings (learned, not sinusoidal) before being concatenated and fed into the transformer.

\paragraph{Masking strategy.}
During training, the mask for $f_2$ is sampled uniformly at random on each example, with exactly 10\% of patches revealed. This ensures the model encounters diverse masking patterns and cannot rely on fixed spatial positions. The asymmetric structure---fully visible $f_1$, 90\% masked $f_2$---is a core design choice that encourages temporal factorization of appearance and motion.

\paragraph{Symmetric masking ablation.}
To test whether this asymmetry is necessary, we trained BabyZWM variants with symmetric masking policies:
\begin{itemize}
    \item \textbf{Symmetric 45\%-45\%}: Both frames are masked at 45\%, so each frame reveals 55\% of patches.
    \item \textbf{Symmetric 90\%-90\%}: Both frames are masked at 90\%, so each frame reveals only 10\% of patches.
\end{itemize}
Both symmetric variants perform substantially worse across all zero-shot visual-cognitive tasks (Figures~\ref{fig:main_flow_depth} and~\ref{fig:main_objective_readout}), demonstrating that the temporally-biased mask structure---rather than masking per se---is critical for learning representations that support flexible zero-shot extraction.

\paragraph{Output and loss.}
The model outputs a prediction $\widehat{f}_2$ for the full second frame, including all masked positions. The training objective is the mean squared error (MSE) between the predicted and ground-truth pixel values of $f_2$, computed over masked patches:
\begin{equation}
    \mathcal{L} = \left\langle \lVert f_2 - \widehat{f}_2 \rVert^{2} \right\rangle_{(f_1, f_2) \in \mathcal{D}}.
\end{equation}

\begin{table}
    \centering
    \caption{\textbf{Model architecture configurations.}
        Architectural hyperparameters for the two ZWM model sizes evaluated in this work.}
    \label{tab:architecture}
    \begin{tabular}{lcc}
        \\
        \hline
        & ZWM-170M & ZWM-1B \\
        \hline
        Transformer layers & 24 & 48 \\
        Attention heads & 12 & 16 \\
        Embedding dimension & 768 & 1280 \\
        Patch size & $8 \times 8$ & $8 \times 8$ \\
        Input resolution & $256 \times 256$ & $256 \times 256$ \\
        Tokens per frame & 1024 & 1024 \\
        Total parameters & $\sim$170M & $\sim$1B \\
        \hline
    \end{tabular}
\end{table}

\subsection*{Training procedure}

Each ZWM model is trained for 200,000 steps with a batch size of 512 (Table~\ref{tab:training}). As the videos are stored at 30 frames per second, this corresponds to $\sim$950 video hours, or roughly 95 days of waking experience assuming $\sim$10 awake hours per day for young children~\cite{iglowstein_sleep_2003}.

Training datapoints consist of RGB frame pairs sampled from real-world video, with the inter-frame temporal gap randomly and uniformly chosen in the range 150--450ms (corresponding to 5--14 frames at 30 fps).

\paragraph{Optimization.}
We use AdamW with a peak learning rate of 3e-4, weight decay of 1e-1, and $(\beta_1, \beta_2) = (0.9, 0.95)$. The learning rate follows a cosine decay schedule with 2000 warmup steps. We use gradient clipping with a maximum norm of 1.0.

\paragraph{Data augmentation.}
No data augmentation (e.g., random crops, color jitter, horizontal flips) is applied during training. The model is trained directly on raw RGB frame pairs.

\paragraph{Compute.}
All models are trained using PyTorch with Distributed Data Parallel (DDP) and mixed-precision (bfloat16). The ZWM-170M model is trained on 4 nodes of 8 NVIDIA H100 GPUs each (32 GPUs total) for approximately 11 hours ($\sim$352 GPU-hours). The ZWM-1B model is trained on 8 nodes of 8 H100 GPUs each (64 GPUs total) for approximately 24 hours ($\sim$1,536 GPU-hours).

\begin{table}
    \centering
    \caption{\textbf{Training hyperparameters.}
        Training configuration shared across all ZWM models unless otherwise noted.}
    \label{tab:training}
    \begin{tabular}{lc}
        \\
        \hline
        Hyperparameter & Value \\
        \hline
        Optimizer & AdamW \\
		AdamW $(\beta_1, \beta_2)$ & (0.9, 0.95) \\
        Peak learning rate & 3e-4 \\
        Weight decay & 1e-1 \\
        LR schedule & Cosine decay \\
        Warmup steps & 2000 \\
        Batch size & 512 \\
        Total training steps & 200,000 \\
        Inter-frame gap & 150--450 ms \\
        Patches revealed during training ($f_1$) & 100\% \\
        Patches revealed during training ($f_2$) & 10\% \\
        \hline
    \end{tabular}
\end{table}

\subsection*{Training datasets}
\label{section:datasets}

We train ZWM on a spectrum of visual diets to test data efficiency and robustness (Table~\ref{tab:datasets}):

\paragraph{BabyView.}
The BabyView dataset~\cite{long_babyview_2024} consists of 868 hours, mostly longitudinal, egocentric video recordings from $N=34$ children aged $\sim$5 months to 3 years, and $\sim$100 hours from 3--5-year-olds recorded in a preschool setting. Videos are recorded using head-mounted cameras worn by the children during natural daily activities. We refer to the ZWM model trained on the full BabyView dataset as ``BabyZWM.''
The raw BabyView videos, recorded by families in their homes, are typically several minutes in duration. We preprocess all videos by splitting them into 10-second clips, stored at 30 fps with the shorter spatial dimension resized to 256 pixels (the longer side scaled proportionally to preserve aspect ratio). During training, we randomly sample a $256 \times 256$ crop from each frame pair, with the same crop applied to both $f_1$ and $f_2$. No additional augmentations are applied.

\paragraph{Single-Child BabyView.}
To test learning from even more restricted experience, we construct a subset of BabyView consisting of 132 hours of recordings from a single individual (child S00320001, aged 9-30 months). This represents the most stringent test of data efficiency, requiring the model to learn generalizable capacities from the highly restricted visual diversity of one child's experience.

\paragraph{Random 132-hour subset.}
To disentangle the contributions of visual diversity from total exposure, we also train on a random 132-hour subset of BabyView, sampled uniformly across all 34 children. This subset matches the Single-Child dataset in total duration but contains substantially more environmental diversity.

\paragraph{Age-ordered curricula.}
To test continual learning and robustness to catastrophic forgetting, we train Single-Child BabyZWM in an online, single-epoch fashion on the age-ordered video stream. We create curricula by shuffling within temporal windows of varying durations:
\begin{itemize}
    \item \textbf{5-minute shuffle}: 10-second clips are randomly shuffled within contiguous 5-minute windows, preserving the coarse temporal order while introducing local mixing (analogous to within-episode consolidation).
    \item \textbf{30-minute shuffle}: 10-second clips are shuffled within 30-minute windows.
    \item \textbf{1-day shuffle}: 10-second clips are shuffled within full recording days, loosely approximating overnight sleep-like reordering.
\end{itemize}

\paragraph{Kinetics-400.}
Kinetics-400~\cite{kay_kinetics_2017} consists of $\sim$670 hours of 10-second video clips from YouTube, spanning 400 human action categories. This dataset is smaller than BabyView but contains substantially more environmental and semantic diversity due to its Internet-sourced content.

\paragraph{Big Video Dataset (BVD).}
BVD~\cite{kotar_world_2025} consists of $\sim$7,000 hours of video drawn from a combination of computer vision datasets and Internet videos. This serves as an approximate upper bound on performance achievable with high visual diversity and scale.

\begin{table}
    \centering
    \caption{\textbf{Training dataset statistics.}
        Summary of the video datasets used to train ZWM variants.}
    \label{tab:datasets}
    \begin{tabular}{lccc}
        \\
        \hline
        Dataset & Hours & Source & ZWM variant \\
        \hline
        BabyView & 868 & 34 children, egocentric & BabyZWM \\
        Single-Child BabyView & 132 & 1 child, egocentric & Single-Child BabyZWM \\
        Random 132h subset & 132 & 34 children, egocentric & --- \\
        Kinetics-400 & $\sim$670 & Internet (YouTube) & ZWM-Kinetics \\
        Big Video Dataset (BVD) & $\sim$7,000 & CV datasets + Internet & ZWM-BVD \\
        \hline
    \end{tabular}
\end{table}

\subsection*{Zero-shot prompt design}
\label{section:specific_prompts}

Here, we describe how diverse visual-cognitive quantities are extracted using ZWM's zero-shot prompts, which act as approximate \textit{causal inferences} by comparing hypothetical or counterfactual predictions against the ground-truth. 
Each prompt follows a common structure: (i) construct a minimal \textbf{perturbation} that intervenes on a latent \textit{cause} governing some visual quantity; (ii) \textbf{compare} the predictor's output under the perturbation against the unperturbed ground-truth prediction; and (iii) \textbf{aggregate} the difference to extract the quantity of interest. 
Simple prompts compose to extract increasingly complex visual structures, building a computational graph of visual intermediates. Table~\ref{tab:prompts} summarizes the structure of each prompt.

\begin{table}
    \centering
    \caption{\textbf{Summary of zero-shot prompts for visual-cognitive tasks.}
        Each prompt extracts a visual quantity by perturbing the predictor's input, comparing the perturbed output to the unperturbed prediction, and aggregating the difference. Later prompts compose earlier ones.}
    \label{tab:prompts}
    \small
    \begin{tabular}{p{0.12\textwidth} p{0.2\textwidth} p{0.22\textwidth} p{0.22\textwidth} p{0.14\textwidth}}
        \\
        \hline
        \textbf{Task} & \textbf{Perturb} & \textbf{Compare} & \textbf{Aggregate} & \textbf{Composes} \\
        \hline
        Optical flow & Add Gaussian tracer to query point in $f_1$ & RGB difference between perturbed and clean predictions of $f_2$ & Argmax of difference gives flow vector & --- (primitive) \\[6pt]
        Hyp.\ motion & Displace object patch to new location in $f_2^{\text{masked}}$ & Predict remaining masked regions & Full hypothetical scene & --- (primitive) \\[6pt]
        Relative depth & Apply optical flow tracer to query point in binocular $f_L$ & Optical flow from $f_L$ to $f_R$ via perturbed vs.\ clean predictions & Rank points by disparity magnitude & Optical flow \\[6pt]
        Object segment.\ & Displace object patch via hyp.\ motion & Optical flow between original and hypothetical scene & Threshold flow; aggregate over directions & Hyp.\ motion, optical flow \\[6pt]
        Intuitive physics & Reveal hand's ground-truth location in $f_2^{\text{masked}}$; fix background & MSE and LPIPS between prediction and target $f_2$ & Closer to target or context; compose flow + segments for ``what moved'' & Hyp.\ motion, optical flow, object segment. \\
        \hline
    \end{tabular}
\end{table}

\textbf{Optical flow} (Figure~\ref{fig:main_flow_depth}A). \textit{Latent cause}: Between two frames $(f_1, f_2)$, a point at position $x_q$ in $f_1$ \textit{causes} a corresponding point in $f_2$, due to the underlying causal structure of motion.
\begin{enumerate}
    \item \textbf{Perturb}: Duplicate the initial frame $f_1$ and add a white-dot tracer to form $\tilde{f}_1$, using a Gaussian centered at the query location $x_q$ with amplitude 255 on each RGB channel and standard deviation $\sigma = 3.0$ pixels.
    \item \textbf{Compare}: Run the model twice with the same masked second frame $f_2^{\text{masked}}$: once with the clean frame $(f_1, f_2^{\text{masked}}) \rightarrow \hat{f}_2$ and once with the perturbed frame $(\tilde{f}_1, f_2^{\text{masked}}) \rightarrow \tilde{f}_2^{\text{pred}}$. Compute the RGB difference $\Delta = \tilde{f}_2^{\text{pred}} - \hat{f}_2$.
    \item \textbf{Aggregate}: Take the argmax of $|\Delta|$ to identify where the perturbation was carried to. The flow vector at $x_q$ is $\text{argmax}(|\Delta|) - x_q$.
\end{enumerate}
Optical flow is the most primitive prompt and does not compose from other prompts; all subsequent prompts build on it. The masked patches in $f_2^{\text{masked}}$ are randomly selected and differ across evaluations, but are held fixed between the perturbed and unperturbed forward passes to ensure the flow signal can be localized.

\textbf{Relative depth} (Figure~\ref{fig:main_flow_depth}D). \textit{Latent cause}: The depth of a point is the latent cause governing its displacement under binocular separation; farther points exhibit smaller binocular disparity.
\begin{enumerate}
    \item \textbf{Perturb}: Given a binocular image pair $(f_L, f_R)$ from stereo cameras, apply the optical flow tracer (as above) to a query point $x_q$ in $f_L$. Note that binocular image pairs are provided by the evaluation dataset; this is ecologically plausible, as humans possess binocular vision.
    \item \textbf{Compare}: Compute the optical flow from $f_L$ to $f_R$ at $x_q$ by comparing the perturbed and unperturbed predictions, composing the optical flow prompt described above.
    \item \textbf{Aggregate}: The magnitude of the resulting flow vector gives the binocular disparity at $x_q$, which is inversely related to depth. To compare relative depth of multiple points, compose multiple optical flow prompts and rank by disparity magnitude.
\end{enumerate}

\textit{Ecological validity and the rigidity question.} The relative-depth readout uses a binocular (stereo) image pair, which humans with two eyes have access to, so a binocular input is ecologically natural. Stereopsis is widespread among animals with overlapping visual fields, and human infants are sensitive to binocular disparity within the first few months \cite{held_stereoacuity_1980}. This readout also does not assume that scenes are rigid. The model uses binocular disparity at each timepoint independently, rather than motion parallax over time, so there is no scene-rigidity-over-time assumption. The two stereo views are captured at the same instant---simulating two eyes---so the disparity between them arises purely from the binocular baseline (parallax) and encodes depth by correspondence regardless of whether objects in the scene are moving. Both static and moving objects have a well-defined stereo disparity.

\textbf{Hypothetical motion} (Figure~\ref{fig:main_segmentation}G). Before describing the remaining prompts, we introduce a key primitive: hypothetical motion. ZWM simulates ``what if this object moved?'' by selecting one or more patches on an object and displacing them to a new location in $f_2^{\text{masked}}$, then predicting the remaining masked regions. The predictor propagates this local displacement to the rest of the object, producing a full hypothetical scene. While displacing a single patch often suffices, displacing multiple patches from the same object generally produces more coherent hypothetical scenes. This primitive is not evaluated directly but serves as a building block for object segmentation and intuitive physics below.

\textbf{Object segmentation} (Figure~\ref{fig:main_segmentation}G). \textit{Latent cause}: Groups of pixels move together due to the latent cause of belonging to the same physical object---a learned form of ``common fate''~\cite{kellman_perception_1983}.
\begin{enumerate}
    \item \textbf{Perturb}: Select a patch on a candidate object and displace it using the hypothetical motion primitive (above), producing a hypothetical scene $\tilde{f}$ in which the object has moved.
    \item \textbf{Compare}: Compose the optical flow prompt to compute flow between the original image $f$ and the hypothetical prediction $\tilde{f}$. Pixels belonging to the perturbed object will exhibit coherent flow; other pixels will not.
    \item \textbf{Aggregate}: Threshold the flow magnitude to produce a binary mask. Repeat over 8 displacement directions, with displacement magnitudes between 25 and 35 pixels, then aggregate the resulting masks to obtain the full object segment.
\end{enumerate}

\textit{Signalling a fixed camera.} During training, the model learns to distinguish camera motion from object motion: when the camera moves, the background and the objects in the scene move together in a coherent manner, whereas when an individual object moves, it moves differently from the background and from other objects. At inference, the benchmark indicates to the model that the camera is not moving by holding the image corners at their original, unmoved appearance. The procedure therefore reveals only (i) the displaced object patch and (ii) these corner blocks. The corners are fixed locations rather than chosen per object; a corner is removed only if it lies close to the displaced source or destination block. This anchoring matters because, without it, an imposed perturbation could itself be read as camera motion, with the whole scene predicted to shift together and the object-specific signal lost. Empirically the anchoring works: in the model's predictions the surrounding background does not change, confirming that the model interprets the grounding patches as signalling a fixed camera, so the resulting segment reflects the imposed object motion rather than camera motion.

Signalling a fixed camera also operationalizes how a human observer might perform this segmentation: faced with a scene, the observer runs the same hypothetical simulation---``what if I nudged this part?''---from a fixed vantage point. Holding the camera fixed is a natural condition for that mental simulation, and it applies equally to static scenes and to scenes containing moving objects. Predicting a hypothetical conditioned on a static camera relies on a signal that biological organisms already possess, since organisms can tell whether their head is moving using efference copy and vestibular input.

\textbf{Intuitive physics} (Figure~\ref{fig:main_object_intuitive_physics}A). \textit{Latent cause}: Physical interactions transmit forces between objects---e.g., pushing one object into another \textit{causes} the second to move, exposing the underlying causal structure of contact dynamics.
\begin{enumerate}
    \item \textbf{Perturb}: Reveal a $32 \times 32$-pixel green intervention patch in $f_2^{\text{masked}}$ at the hand's ground-truth location in the target frame, providing information about where the hand has moved. The hand location is annotated by human labelers, with care taken to ensure the patch does not reveal the object's position. This acts as a perturbation relative to the unperturbed case (where the hand's motion is masked and unknown), prompting the model to predict the physical consequences of the hand's action on the rest of the scene. Additional $32 \times 32$-pixel red background patches are revealed to fix illumination and camera pose, isolating the causal effect of the hand's motion on objects. These border patches are revealed at their true, unchanged appearance to signal a static camera. The model operates directly on the revealed pixels, so it has access to their fine detail; empirically its predictions keep the table and surrounding background fixed, confirming that it reads the grounding patches as a stationary camera.
    \item \textbf{Compare}: Compare the model's prediction (given the revealed hand motion) against the ground-truth target frame $f_2$, using both MSE and LPIPS perceptual similarity~\cite{zhang_unreasonable_2018}.
    \item \textbf{Aggregate}: Determine whether the prediction is closer to the ground-truth target $f_2$ or the context frame $f_1$. Additionally, compose optical flow and object segmentation prompts on the predicted scene to evaluate \textit{what} moved and \textit{how}---e.g., whether force transferred to a second object.
\end{enumerate}

\subsection*{Evaluation benchmarks}
\label{section:evaluation}

\paragraph{Optical flow: TAP-Vid benchmarks.}
We evaluate optical flow on two benchmarks from the TAP-Vid suite:
\begin{itemize}
    \item \textbf{TAP-Vid-DAVIS}~\cite{doersch_tap-vid_2023}: Real-world videos with human-annotated ground-truth point correspondences, featuring challenging scenarios including fast motion, occlusions, and appearance changes.
    \item \textbf{TAP-Vid-Kubric}~\cite{greff_kubric_2022}: Synthetic, simulator-generated videos where ground-truth flows are known by construction, providing a complementary evaluation without annotation noise.
\end{itemize}
\textit{Query protocol and temporal-gap distribution.} We use TAP-Vid's standard ``first'' query mode, in which every visible point is matched from its first visible occurrence to every subsequent target frame. To enable a controlled comparison across model classes, every model receives exactly two frames---the query frame and the target frame---and nothing else; this is the native input format for ZWM and for the two-frame optical-flow models. The temporal gap between those two frames is therefore not a parameter we choose, but the distribution induced by the benchmark's own query mode. Across the 38{,}881 query--target pairs of TAP-Vid-DAVIS (650 query points over 30 videos), the gap ranges from 0 to 103 frames, with a median of 31 and a mean of 34. The distribution spans the full range rather than concentrating at a single stride (Table~\ref{tab:tapvid_gaps}), so the benchmark tests correspondence across both short gaps and large displacements.

\begin{table}[H]
    \centering
    \caption{\textbf{Temporal-gap distribution of the TAP-Vid-DAVIS ``first'' query protocol.}
        Share of the 38{,}881 query--target pairs falling in each gap range. The distribution is induced by the benchmark's query mode rather than chosen: every visible point is matched from its first visible occurrence to every subsequent frame.}
    \label{tab:tapvid_gaps}
    \begin{tabular}{lc}
        \\
        \hline
        Temporal gap (frames) & Share of pairs \\
        \hline
        0--10   & 18.2\% \\
        11--30  & 31.3\% \\
        31--60  & 34.4\% \\
        $>$60   & 16.1\% \\
        \hline
        Median 31; mean 34; range 0--103 & 38{,}881 pairs \\
        \hline
    \end{tabular}
\end{table}

\textit{Optical flow versus tracking.} These two literatures use different input regimes. Optical-flow methods estimate motion from a pair of frames, whereas point-tracking methods consume a whole video and chain predictions across it. Although we evaluate in a two-frame optical-flow setting, we use the TAP-Vid tracking benchmark because it contains real-world, challenging dynamics, whereas optical-flow benchmarks are mostly synthetically generated (e.g., Sintel, FlyingThings). Passing every model the same two frames therefore lets us compare across both model classes on the same input, and this pairwise implementation of the TAP-Vid ``first'' protocol has been used in prior work~\cite{doersch_tap-vid_2023, jiang_doduo_2023, guerrier_pointst3r_2025}.

\textit{Why two frames.} Two-frame optical flow is a classic and important scientific concept~\cite{gibson_optical_1957}, and historically it played a key role in thinking about visual development. Two-frame evaluation is the fundamental unit of motion understanding --- tracking itself is built from estimating how a point moves or becomes occluded across two frames --- and it isolates the underlying motion representation by removing algorithm-specific heuristics for chaining frame-pair predictions into tracks. It is the standard regime of the optical-flow community, whereas multi-frame evaluation is standard in the tracking community; the two simply reflect different goals.

\textit{Tracking algorithms and temporal-smoothness priors.} Tracking algorithms depend on temporal-smoothness priors, so the tracking evaluation --- although useful for computer-vision applications --- is less indicative of a model's capacity to understand the core motion-dynamics computation we are concerned with. These models build in many assumptions about smoothness, which is orthogonal to the vision-science question. Precisely because the multi-frame numbers are naturally higher, they allow an important dependence on smoothness priors that can blunt the assessment of how good a model's understanding of basic dynamics is. The two-frame CoTracker3 results are nonetheless reasonable: its performance is degraded reasonably rather than fully, still matching the supervised flow models DPFlow and SeaRAFT, which makes sense because all three take in two frames; and multi-frame CoTracker3, trained on multiple labelled datasets, achieves similar performance to ZWM-1B, which reaches 74.0\% while seeing only two frames. We read this not as a deficiency of the two-frame protocol but as evidence that much of a tracker's advantage comes from aggregating intermediate frames rather than from stronger core two-frame dynamics understanding.

\textit{Downstream relevance.} We are primarily interested in optical flow as a scientific concept and an instantiation of a concrete visual ability, rather than as a computer-vision application. Nevertheless, two-frame correspondence is the basic primitive behind frame interpolation, video compression, and motion-based segmentation --- settings where motion must be estimated from a pair of frames rather than from a long continuous clip. We view two-frame flow and multi-frame tracking as complementary rather than opposed: algorithms that are good at two-frame flow can be made into tracking algorithms, and historically advances in optical flow have driven better multi-frame trackers~\cite{jiang_doduo_2023}.

All evaluations are conducted at $256 \times 256$ resolution. For each algorithm, we report two standard TAP-Vid metrics~\cite{doersch_tap-vid_2023}:
\begin{itemize}
    \item \textbf{Position accuracy ($< \delta^x_{\text{avg}}$)}: For visible points, the fraction of predicted correspondences falling within a pixel-distance threshold of the ground-truth position, averaged over five thresholds (1, 2, 4, 8, and 16 pixels).
    \item \textbf{Occlusion accuracy (OA)}: Binary classification accuracy for predicting whether each query point is occluded or out of frame on each time step.
\end{itemize}

\paragraph{Relative depth: UniQA-3D.}
We evaluate relative depth estimation on UniQA-3D~\cite{zuo_towards_2024}, which presents pairs of points and requires judging which is farther from the camera.
The upright data originally contains 500 samples, but after filtering to ensure the query points fall unambiguously within the center crop of the image and that each image contains a stereo pair from the original KITTI dataset, the upright set is filtered to 103 examples and the flipped set to 61 examples.

\paragraph{Object segmentation: SpelkeBench.}
We evaluate class-agnostic object segmentation on SpelkeBench~\cite{venkatesh_discovering_2025}, which defines objects as distinct, bounded physical entities. The benchmark draws images from two sources: 496 images from EntitySeg (real-world scenes)~\cite{qi_high-quality_2023} and 51 images from Open X-Embodiment (real-world robot interactions)~\cite{collaboration_open_2025}, totaling 547 images. We measure performance using intersection-over-union (IoU)~\cite{lin_microsoft_2015}.

\paragraph{Intuitive physics benchmark.}
We develop a novel short-timescale physical reasoning benchmark to evaluate models on intuitive physics (Figure~\ref{fig:main_object_intuitive_physics}A). The benchmark features tabletop interactions between a hand and 1--2 objects, testing five categories of physical reasoning:
\begin{enumerate}
    \item \textbf{Object cohesion}: When one part of an object is moved, the entire object moves together.
    \item \textbf{Support (top object moves)}: When a supporting surface is removed, the supported object falls.
    \item \textbf{Support (bottom object moves)}: When the bottom object in a stack is moved, the top object moves with it.
    \item \textbf{Force transfer}: Pushing one object into another causes the second object to move.
    \item \textbf{Force separation}: Moving one object does not affect a spatially separated object.
\end{enumerate}
The benchmark contains 20 image pairs per category (100 image pairs total). Each image pair is evaluated under 8 different random mask configurations for the revealed patches in $f_2$, yielding $5 \times 20 \times 8 = 800$ total evaluations per model.

We first considered classic object-permanence tasks (e.g., violation-of-expectation under occlusion). But these primarily probe working memory, rather than reasoning about physical dynamics. This leads to unnecessary confounds; if a model fails, it is not obvious whether it is due to limited working memory. Existing object-permanence benchmarks were set up this way because they were inspired by tests on human infants, whereas our intuitive-physics benchmark is designed for models and deliberately short-timescale (two frames), isolating physical reasoning --- cohesion, support, force transfer and force separation --- from the memory demands that object permanence introduces. Tackling object permanence calls for longer-horizon prediction and temporal memory, which we flag in the Discussion as important future work.

Each example consists of a context frame and a target frame. Accuracy is defined as the proportion of examples for which the model's prediction is closer to the ground-truth target than to the context frame, evaluated using both MSE and LPIPS~\cite{zhang_unreasonable_2018}.

\textit{Accuracy, stated precisely.} For each example, the model's prediction is counted as correct if it is closer---that is, at a smaller distance---to the ground-truth target frame than to the context frame. This is computed under two distance measures, mean squared error (MSE) and LPIPS perceptual similarity~\cite{zhang_unreasonable_2018}, run separately, yielding two accuracy numbers per category. MSE and LPIPS capture different things---raw pixel error versus learned perceptual similarity---yet accuracy is very similar under both, which adds robustness, since the result does not hinge on a particular choice of distance measure.

\textit{A graded alternative to the binary metric.} Because this binary accuracy saturates near ceiling and gives only a coarse pass/fail picture, we additionally report a finer, graded measure of how faithfully each prediction matches the real outcome. It is a continuous closeness score---the fraction of the context-to-target gap that the prediction closes,
\begin{equation}
    \text{graded} = \frac{d(\text{pred}, \text{context})}{d(\text{pred}, \text{context}) + d(\text{pred}, \text{target})} \in [0, 1],
\end{equation}
where a value of 0 means the prediction is identical to the context (no change predicted) and a value of 1 means it is identical to the ground-truth outcome. We compute $d$ under both the MSE and the optical-flow distances, and within each object region (primary, secondary, and overall), turning the binary accuracy into a continuous fidelity scale. These graded scores rank the models exactly as the binary accuracy does, confirming the original result, but resolve differences that the binary number hides (Tables~\ref{tab:phys_motion_fidelity} and~\ref{tab:phys_graded}).

\textit{Scoring a latent-prediction model.} V-JEPA2 does not generate pixels, so we do not force it into a pixel metric. It is evaluated in its native feature space --- we compare whether its predicted features are closer to the target frame's features than to the context frame's --- which is the appropriate way to score a latent-prediction model, and prior work has used V-JEPA2 to predict and compare in embedding space, including the original V-JEPA2 paper~\cite{assran_v-jepa_2025}. Given that some models predict pixels whereas others predict latent representations, the fairest approach is to evaluate each model in its native prediction space. The prediction spaces are nonetheless different and not directly comparable, so the feature-based and pixel-based columns of Table~\ref{tab:phys_graded} should not be read against one another.

\paragraph{Developmental trajectories.}
To analyze developmental curves, we evaluate BabyZWM at various training checkpoints (0, 5k, 10k, 20k, 40k, 80k, 120k, 160k, 200k). Each ZWM model is trained for 200,000 steps with a batch size of 512. As the videos are stored at 30 frames per second, this corresponds to $\sim$950 video hours, or roughly 95 days of waking experience assuming $\sim$10 awake hours per day for young children~\cite{iglowstein_sleep_2003}. The $x$-axis represents training steps.

\paragraph{Neural predictivity.}
We evaluate the alignment between model representations and biological neural responses using two complementary benchmarks:
\begin{itemize}
    \item \textbf{Natural Scenes Dataset (NSD)}~\cite{allen_massive_2022}: Human fMRI responses to natural images, capturing large-scale representational geometry.
    \item \textbf{THINGS Ventral Stream Spiking Dataset (TVSD)}~\cite{papale_extensive_2025}: Macaque single-neuron electrophysiology, revealing fine-grained neural tuning and timing.
\end{itemize}
For each model and brain region, we:
\begin{enumerate}
    \item Extract features from every other layer of the model for each stimulus image.
    \item Fit a cross-validated ridge regression from model features to neural responses.
    \item Report noise-corrected Pearson correlations as the measure of neural predictivity.
\end{enumerate}
For both NSD and TVSD, we use 10-fold cross-validation to split the data into training and test sets. Ridge regression regularization is performed using \texttt{RidgeCV}, which evaluates 21 regularization strengths ($\alpha$). Importantly, $\alpha$ is selected independently for each target (i.e., per voxel for fMRI, per neuron for electrophysiology), allowing the regularization to adapt to the noise characteristics of each recording site.

Noise ceilings are estimated differently for each dataset. For NSD, we use the reliability estimation method described by Allen et al.~\cite{allen_massive_2022}. For TVSD, noise ceilings are computed via split-half correlations. For NSD, neural predictivity is evaluated for V1, V2, V3, V4, and the mid- and high-ventral visual regions. For TVSD, we evaluate using V1, V4, and inferior temporal (IT) cortex.

\subsection*{Baselines}
\label{section:baselines}

We compare ZWM against both representation-based models and task-specific systems.

\paragraph{Representation-based models.}
Unlike ZWM, representation-based models are not natively zero-shot and typically require labeled supervision (fine-tuning or linear probes) for each downstream task. To enable fair comparison, we design simple zero-shot probes for these models:

\begin{itemize}
    \item \textbf{ResNet50 (ImageNet-supervised)}: A standard ResNet50~\cite{he_deep_2015} pretrained on ImageNet-1K with category-label supervision.

    \item \textbf{Baby DINOv3}: DINOv3~\cite{simeoni_dinov3_2025} (ViT-Large) learns single-image representations by training the model to produce consistent features across different augmented views of the same image. We train DINOv3 on BabyView (868 hours).

    \item \textbf{Baby V-JEPA2}: V-JEPA2 is a self-supervised video model that learns by predicting masked regions of a video in feature space rather than in raw pixels. We train a 300-million parameter V-JEPA2 model on BabyView (868 hours) using the official implementation and default hyperparameters. We verified successful training via frozen linear probes on held-out subsets, yielding 54.2\% top-1 accuracy on Kinetics-400 (400-way classification) and 53.45\% on ImageNet-1K (1000-way classification)~\cite{deng_imagenet_2009}.
\end{itemize}

In addition to these BabyView-trained variants, we evaluate the publicly released, standard-data-trained DINOv3 (ViT-Large) and V-JEPA2 (Large and Giant) models as non-developmental reference points; these are the ``standard'' / non-``Baby'' rows in the supplementary comparison tables.

\paragraph{Zero-shot probe designs for representation-based baselines.}
Since ResNet50, DINOv3 and V-JEPA2 are representation-based models that do not natively support zero-shot visual-cognitive extraction, we design simple probe procedures to enable fair comparison.

\textit{Optical flow.}
For ResNet50, DINOv3 and V-JEPA2, we pass both frames through the model and extract patch-level feature representations. To estimate the flow at a query point in the first frame, we compute the cosine similarity between the query point's patch feature in the first frame and all patch features in the second frame. The target location is taken as the patch in the second frame with the highest cosine similarity, and the flow vector is the displacement between the query and target positions. This cosine-similarity nearest-neighbor matching follows the protocol of the DINOv3 paper~\cite{simeoni_dinov3_2025} and is a standard label-free readout in prior self-supervised work~\cite{caron_emerging_2021, gupta_siamese_2023}; we additionally apply multi-iteration zooming to raise the baselines' effective feature resolution.

\textit{Relative depth.}
We apply the same cosine-similarity correspondence matching procedure described above for optical flow to binocular image pairs, and infer relative depth from the magnitude of the resulting disparity (optical flow) vector, following the same logic as the ZWM depth prompt.

\textit{Object segmentation.}
We compute pairwise cosine similarity between all patch features within each image. Object segments are obtained by thresholding the cosine similarity to a seed patch, grouping patches with high feature similarity as belonging to the same object.

\textit{Intuitive physics.}
ResNet50 and DINOv3 are single-image models and cannot be meaningfully evaluated on our temporal intuitive physics benchmark, so they are excluded from this comparison. For V-JEPA2, we provide the same revealed patches (hand location and background grounding patches) as input and use V-JEPA2 to predict representations for the masked patches. We then compute the cosine similarity between the predicted representations and the representations of the same regions extracted from (i) the initial context frame $f_1$ and (ii) the ground-truth target frame $f_2$. If the predicted representations are more similar to $f_2$ than to $f_1$, the model is scored as correct.

\paragraph{Task-specific baselines.}
For each visual-cognitive task, we compare against state-of-the-art task-specific models:

\begin{itemize}
    \item \textbf{Optical flow}: CoTracker3~\cite{karaev_cotracker_2024}, DPFlow~\cite{morimitsu_dpflow_2025}, and SeaRAFT~\cite{wang_sea-raft_2024}. All are supervised models trained with ground-truth flow annotations.

    \item \textbf{Relative depth}: MiDaS-CNN (supervised monocular)~\cite{ranftl_towards_2020}, MonoDepth2 (self-supervised monocular)~\cite{godard_digging_2019}, ProDepth (self-supervised monocular)~\cite{woo_prodepth_2024}, and FoundationStereo (supervised binocular)~\cite{wen_foundationstereo_2025}. We also compare against large vision-language models: Gemini-1.5~\cite{gemini_gemini_2024}, GPT-4-Turbo~\cite{openai_gpt-4_2024}, and GPT-4o~\cite{openai_gpt-4o_2024}.

    \item \textbf{Object segmentation}: Mask2Former~\cite{cheng_masked-attention_2022} (trained on COCO~\cite{lin_microsoft_2015}) and SAM2~\cite{ravi_sam_2024} (trained with large-scale human annotations).

    \item \textbf{Intuitive physics}: No established baselines exist for our novel benchmark; we compare against V-JEPA2 and Baby V-JEPA2.
\end{itemize}

\newpage


\section*{Supplementary Text}

\subsection*{Relation to prior work on frame-pair masked prediction}

We separate ZWM's core idea into two components: (i) the representation-learning objective (temporally-factored masked prediction), and (ii) how that representation is used to read out visual structure. On (i) we overlap with prior masked auto-encoding work, e.g., SiamMAE \cite{gupta_siamese_2023}. Our key contribution is (ii), a zero-shot causal-inference mechanism that reads flow, depth, segmentation, and intuitive physics out of the frozen predictor.

ZWM's zero-shot extraction mechanism is different from prior work; it uses approximate causal inference that can be further composed toward increasingly complex abilities. We perturb the input, run the predictor, and compare the perturbed prediction against the unperturbed one, exposing the causal structure of the world (for instance, pixels move together because they belong to the same physical object). This perturb--predict--compare is generic and unifies the extraction of a broad, composable set of quantities --- optical flow, relative depth, object segmentation. Feature-propagation approaches such as SiamMAE \cite{gupta_siamese_2023} show neither depth nor intuitive-physics tests, nor a compositional interface.

Furthermore, at its time of release, SiamMAE \cite{gupta_siamese_2023} was the strongest self-supervised correspondence learner on propagation-style benchmarks, but well below task-specialized supervised state-of-the-art. On optical flow specifically, SiamMAE did not evaluate head-to-head against flow methods. By contrast, ZWM is competitive with supervised, task-specialized state-of-the-art across optical flow, relative depth, and segmentation.

Empirically, ZWM's zero-shot causal-inference readout is an important source of performance. Read with feature-propagation, the standard BabyZWM-170M model scores 0.526 on optical flow, 87.2\% on relative depth, and 0.417 mIoU on segmentation. Read with approximate causal inference, it achieves 0.616, 94.2\%, and 0.496 (Table~\ref{tab:mask_ablation}; main-text Figure~\ref{fig:main_objective_readout}). Hence, applying a SiamMAE-style feature-propagation readout to ZWM recovers substantially less of the representation than our mechanism does, and the matched-probe comparisons in the following subsections show the same pattern across every model we evaluate. The full crossing of training objective with readout mechanism, including the symmetric-masking variants, is reported below under ``Ablation: training objective $\times$ readout mechanism''.

\subsection*{What we mean by ``zero-shot''}

By ``zero-shot'' we mean the model performs a task with no task-specific training and examples.

We wish to highlight an important distinction between \emph{examples} and \emph{instructions}. In the LLM literature, models are commonly provided with task instructions (what the model is supposed to do) and examples of the task being completed. ``Zero-shot'' refers to performance when these task-specific examples are omitted. This aligns with the machine learning literature, which contrasts ``few-shot'' against ``zero-shot'' learning. In few-shot learning, several examples are provided to the model, either in context or for updating the weights through learning; in image recognition, for example, few-shot learning refers to providing a few exemplars of a novel, unseen category to a model, and using it to perform image recognition for the category. ``Zero-shot'' is the complement of the term, meaning that no examples are provided. ZWM's readout programs serve as \emph{instructions} --- without examples --- just like prompts for LLMs.

Our usage of the term also aligns with a developmental framing. A child --- whom the model is meant to model --- comes to estimate motion, depth, object segments, and physical outcomes without ever being trained on those tasks or shown labeled examples of them. ZWM mirrors this.

Some task-specific ``domain knowledge'' or ``instructions'' must nonetheless be supplied to any model. This can come in the form of supervised labels, or alternatively as instructions that communicate a procedure for solving a task, exactly what our zero-shot probes are. Otherwise there is no way for a system to know what task to perform and how. Consequently, the nearest-neighbor feature-matching probe we apply to the representation baselines is also a form of task-specific instruction. For flow extraction using feature-NN matching, the specific set of instructions are: pass both frames into the model, build representations across the model, pick a specific layer (the specific layer is an important, task-specific choice), take the query patch's representation, use cosine similarity to find the target patch with the closest representation. All these steps are important innate elements provided to the model, and they are task-specific; the steps are different for performing object segmentation, which instead involves single images and thresholding cosine similarity. In this sense, feature-NN matching encodes domain knowledge, and is similarly ``zero-shot'' as our approximate causal-inference mechanism.

Although we provide innate task-specific structure, our framework of approximate causal inference is generic and unifies several tasks, each using the same high-level recipe. First, construct a minimal perturbation that intervenes on a latent cause governing some visual quantity. Second, compare the prediction under the perturbed input against the prediction under the original input. Third, aggregate the difference to extract the quantity of interest. Flow, depth, segments, and intuitive physics differ only in which latent cause is intervened upon; the overall recipe --- perturb, predict, compare/aggregate --- is unchanged. The developmental plausibility of this built-in structure, and the sense in which the readout programs are hypothesized innate priors rather than learned task heads, are discussed in the main text.

Furthermore, our usage of the term ``zero-shot'' is stricter and more principled than common usage in AI in several respects. Recent computer-vision work uses ``zero-shot'' to mean out-of-distribution generalization: a model is trained on a task with labels, then transferred ``zero-shot'' to the same task on unseen benchmarks \cite{ravi_sam_2024}. Our usage is stricter, since we never train on the task and use no task labels at any point. In the LLM literature, ``zero-shot'' has come to mean flexibly performing a new task from a prompt; but because such models are trained on enormous corpora, many ``zero-shot'' tasks are basically interpolations of the training distribution, and what counts as genuinely ``zero-shot'' is ambiguous. By that measure our setting is again more demanding: BabyZWM learns from a few hundred hours of children's video and is tested for capabilities it was never trained on.

\subsection*{Optical flow: Matched feature-NN probing across models}

We apply the same matched feature-NN probe to every model, sweeping the transformer layer from which the frozen features are read (Table~\ref{tab:flow_sweep}) and reporting each model at its best-performing layer (Table~\ref{tab:flow_methods}).

\begin{table}[H]
    \centering
    \caption{\textbf{Matched feature-NN flow probe: per-layer sweep, all ZWM models.}
        The same nearest-neighbor cosine patch-feature probe used for the representation baselines, applied to each model's own frozen features (in-context two-frame encoding, parameter-free LayerNorm), on TAP-Vid-DAVIS two-frame ``first'' (position accuracy, higher is better). 170M sweeps layers $\{0,4,8,12,16,20,-1\}$; 1B sweeps $\{0,8,16,24,32,40,-1\}$; --- $=$ layer not in this model's sweep grid.}
    \label{tab:flow_sweep}
    \begin{tabular}{lcccc}
        \\
        \hline
        Layer & ZWM-170M & ZWM-1B & BabyZWM-170M & BabyZWM-1B \\
              & (BigVideo) & (BigVideo) & (BabyView) & (BabyView) \\
        \hline
        L0    & 0.191  & 0.142  & 0.222  & 0.159 \\
        L4    & 0.220  & ---    & 0.317  & --- \\
        L8    & 0.240  & 0.179  & 0.410  & 0.315 \\
        L12   & 0.325  & ---    & 0.489  & --- \\
        L16   & 0.423  & 0.303  & 0.526  & 0.414 \\
        L20   & 0.455  & ---    & 0.526  & --- \\
        L24   & ---    & 0.440  & ---    & 0.469 \\
        L32   & ---    & 0.514  & ---    & 0.491 \\
        L40   & ---    & 0.565  & ---    & 0.484 \\
        $-1$  & 0.312  & 0.453  & 0.449  & 0.276 \\
        \hline
    \end{tabular}
\end{table}

\begin{table}[H]
    \centering
    \small
    \caption{\textbf{Optical flow, all methods (TAP-Vid-DAVIS ``first'' position accuracy, Pct).}
        Same 38{,}881-point set; res $=$ input resolution (patch grid). feature-NN $=$ best over the layer sweep (best layer in parens).}
    \label{tab:flow_methods}
    \begin{tabular}{llcc}
        \\
        \hline
        Model & probe & res & Pct \\
        \hline
        \multicolumn{4}{l}{\textit{ZWM: native zero-shot inference}} \\
        ZWM-170M (BigVideo)     & \textbf{native} (perturb-compare) & 256 & 0.713 \\
        ZWM-1B (BigVideo)       & \textbf{native} (perturb-compare) & 256 & \textbf{0.740} \\
        BabyZWM-170M (BabyView) & \textbf{native} (perturb-compare) & 256 & 0.616 \\
        BabyZWM-1B (BabyView)   & \textbf{native} (perturb-compare) & 256 & 0.650 \\
        \hline
        \multicolumn{4}{l}{\textit{feature-NN probe}} \\
        ZWM-170M (BigVideo)     & feature-NN (L20)              & 256 ($32\times32$) & 0.455 \\
        ZWM-1B (BigVideo)       & feature-NN (L40)              & 256 ($32\times32$) & \textbf{0.565} \\
        BabyZWM-170M (BabyView) & feature-NN (L16)             & 256 ($32\times32$) & 0.526 \\
        BabyZWM-1B (BabyView)   & feature-NN (L32) & 256 ($32\times32$) & 0.491 \\
        ResNet50                & feature-NN (final)            & 256          & 0.419 \\
        DINOv3 (Large)     & feature-NN (final)            & 224 ($14\times14$) & 0.375 \\
        Baby DINOv3 (Large)     & feature-NN (final)            & 224 ($14\times14$) & 0.349 \\
        V-JEPA2 (Large)    & feature-NN (final)            & 256 ($16\times16$) & 0.293 \\
        Baby V-JEPA2 (Large)    & feature-NN (final)            & 256 ($16\times16$) & 0.215 \\
        \hline
        \multicolumn{4}{l}{\textit{multi-frame tracker (full video, reference)}} \\
        CoTracker3 (multi-frame) & supervised (full-video track) & full video & 0.745 \\
        \hline
        \multicolumn{4}{l}{\textit{supervised flow / 2-frame (reference)}} \\
        DPFlow              & supervised & 2-frame & \textbf{0.625} \\
        SeaRAFT             & supervised & 2-frame & 0.587 \\
        CoTracker3 (2-frame) & supervised & 2-frame & 0.580 \\
        RAFT                & supervised & 2-frame & 0.544 \\
        Doduo               & supervised & 2-frame & 0.485 \\
        \hline
        \multicolumn{4}{l}{\textit{self-supervised (reference)}} \\
        GMRW      & self-supervised & 2-frame & 0.546 \\
        DiffTrack & self-supervised & 2-frame & 0.469 \\
        SMURF     & self-supervised & 2-frame & 0.442 \\
        \hline
    \end{tabular}
\end{table}

\clearpage
\subsection*{Relative depth: Matched feature-NN probing across models}

The same probe is applied to the binocular stereo pair, again sweeping the feature layer (Table~\ref{tab:depth_sweep}) and reporting each model at its best-performing layer (Table~\ref{tab:depth_methods}).

\begin{table}[H]
    \centering
    \caption{\textbf{Matched feature-NN relative-depth probe: per-layer sweep, all ZWM models (\% accuracy).}
        Cosine patch-feature NN probe applied to the binocular stereo pair (disparity correspondence) on UniQA-3D / KITTI (164 pairs); chance $=$ 50\%. 170M sweeps $\{0,4,8,12,16,20,-1\}$; 1B sweeps $\{0,8,16,24,32,40,-1\}$; --- $=$ layer not in this model's sweep grid.}
    \label{tab:depth_sweep}
    \begin{tabular}{lcccc}
        \\
        \hline
        Layer & ZWM-170M & ZWM-1B & BabyZWM-170M & BabyZWM-1B \\
              & (BigVideo) & (BigVideo) & (BabyView) & (BabyView) \\
        \hline
        L0    & 53.7   & 48.8   & 57.3   & 51.2 \\
        L4    & 54.9   & ---    & 72.0   & --- \\
        L8    & 57.9   & 54.3   & 81.7   & 63.4 \\
        L12   & 67.1   & ---    & 84.8   & --- \\
        L16   & 68.3   & 64.0   & 86.0   & 76.8 \\
        L20   & 74.4   & ---    & 87.2   & --- \\
        L24   & ---    & 69.5   & ---    & 80.5 \\
        L32   & ---    & 79.3   & ---    & 82.3 \\
        L40   & ---    & 90.2   & ---    & 82.3 \\
        $-1$  & 74.4   & 80.5   & 82.3   & 68.9 \\
        \hline
    \end{tabular}
\end{table}

\begin{table}[H]
    \centering
    \small
    \caption{\textbf{Relative depth, all methods (\% accuracy; chance $=$ 50\%).}
        ZWM-native and reference rows are canonical values; feature-NN-on-ZWM rows are the matched runs (best completed layer in parens).}
    \label{tab:depth_methods}
    \begin{tabular}{llc}
        \\
        \hline
        Model & probe & acc \\
        \hline
        \multicolumn{3}{l}{\textit{ZWM: native zero-shot inference}} \\
        ZWM-170M (BigVideo)     & \textbf{native} (perturb-compare) & \textbf{98.1} \\
        ZWM-1B (BigVideo)       & \textbf{native} (perturb-compare) & 96.1 \\
        BabyZWM-170M (BabyView) & \textbf{native} (perturb-compare) & 94.2 \\
        BabyZWM-1B (BabyView)   & \textbf{native} (perturb-compare) & 89.3 \\
        \hline
        \multicolumn{3}{l}{\textit{feature-NN probe: ZWM (ours, new)}} \\
        ZWM-170M (BigVideo)     & feature-NN (L20) & 74.4 \\
        ZWM-1B (BigVideo)       & feature-NN (L40) & \textbf{90.2} \\
        BabyZWM-170M (BabyView) & feature-NN (L20) & 87.2 \\
        BabyZWM-1B (BabyView)   & feature-NN (L40) & 82.3 \\
        \hline
        \multicolumn{3}{l}{\textit{feature-NN probe: representation baselines}} \\
        ResNet50             & feature-NN & 71.8 \\
        DINOv3 (Large)  & feature-NN & 63.1 \\
        Baby DINOv3 (Large)  & feature-NN & 72.8 \\
        V-JEPA2 (Large) & feature-NN & 46.6 \\
        V-JEPA2 (Giant) & feature-NN & 57.3 \\
        Baby V-JEPA2 (Large) & feature-NN & 68.0 \\
        \hline
        \multicolumn{3}{l}{\textit{supervised (reference)}} \\
        FoundationStereo (binocular) & supervised & \textbf{100.0} \\
        MiDaS-DPT (monocular)        & supervised & 89.3 \\
        MiDaS-CNN (monocular)        & supervised & 88.3 \\
        \hline
        \multicolumn{3}{l}{\textit{self-supervised (reference)}} \\
        MonoDepth2 & self-supervised & 94.2 \\
        ProDepth   & self-supervised & 89.3 \\
        \hline
        \multicolumn{3}{l}{\textit{humans / vision-language (reference)}} \\
        Humans (single image) & ---  & 84.5 \\
        GPT-4o                & VLM  & 65.0 \\
        Gemini-1.5            & VLM  & 64.1 \\
        GPT-4-Turbo           & VLM  & 50.5 \\
        \hline
    \end{tabular}
\end{table}

\clearpage
\subsection*{Object segmentation: Matched feature-NN probing across models}

For segmentation the probe has two free choices, the feature layer and the bipartition threshold $\tau$, so we sweep both (Tables~\ref{tab:seg_tau} and~\ref{tab:seg_layermodel}) and report each model at its best-performing configuration (Table~\ref{tab:seg_methods}).

\begin{table}[H]
    \centering
    \small
    \setlength{\tabcolsep}{4pt}
    \caption{\textbf{Matched feature-NN segmentation probe: $\tau$ sweep, ZWM-1B (BigVideo).}
        Cosine patch-feature probe on SpelkeBench (547 images): cosine-affinity from each GT seed patch, threshold at bipartition $\tau$, connected component containing the seed, upsample. Metric: IoU. Rows $=$ feature layer; columns $=$ bipartition $\tau$; \textbf{bold} $=$ best per row.}
    \label{tab:seg_tau}
    \begin{tabular}{lccccccccc}
        \\
        \hline
        Layer & 0.1 & 0.2 & 0.3 & 0.4 & 0.5 & 0.6 & 0.7 & 0.8 & 0.9 \\
        \hline
        L0   & .096 & .096 & .096 & .096 & .097 & .101 & .135 & \textbf{.188} & .180 \\
        L8   & .096 & .096 & .096 & .098 & .109 & .149 & .264 & \textbf{.302} & .148 \\
        L16  & .096 & .103 & .126 & .187 & .314 & \textbf{.385} & .293 & .134 & .045 \\
        L24  & .100 & .120 & .181 & .301 & \textbf{.385} & .346 & .208 & .090 & .037 \\
        L32  & .102 & .123 & .184 & .296 & \textbf{.374} & .328 & .184 & .075 & .033 \\
        L40  & .110 & .145 & .222 & .303 & \textbf{.327} & .267 & .155 & .068 & .032 \\
        L47  & .096 & .097 & .101 & .109 & .121 & .153 & .194 & \textbf{.200} & .089 \\
        $-1$ & .099 & .106 & .119 & .134 & .165 & .207 & \textbf{.240} & .216 & .104 \\
        \hline
    \end{tabular}
\end{table}

\begin{table}[H]
    \centering
    \caption{\textbf{Segmentation feature-NN: layer $\times$ model, best over $\tau$.}
        Each cell $=$ best IoU over $\tau \in \{0.1,\dots,0.9\}$; --- $=$ layer not in this model's sweep grid (170M sweeps $\{0,4,8,12,16,20,23,-1\}$; 1B sweeps $\{0,8,16,24,32,40,47,-1\}$). \textbf{bold} $=$ best layer per model.}
    \label{tab:seg_layermodel}
    \begin{tabular}{lcccc}
        \\
        \hline
        Layer & ZWM-170M & ZWM-1B & BabyZWM-170M & BabyZWM-1B \\
              & (BigVideo) & (BigVideo) & (BabyView) & (BabyView) \\
        \hline
        L0    & .258   & .188   & .295   & .197 \\
        L4    & .268   & ---    & .405   & --- \\
        L8    & .295   & .302   & .403   & .348 \\
        L12   & \textbf{.383} & ---    & .365   & --- \\
        L16   & .377   & \textbf{.385} & \textbf{.417} & .398 \\
        L20   & .360   & ---    & .369   & --- \\
        L23   & .214   & ---    & .238   & --- \\
        L24   & ---    & .385   & ---    & \textbf{.414} \\
        L32   & ---    & .374   & ---    & .384 \\
        L40   & ---    & .327   & ---    & .344 \\
        L47   & ---    & .200   & ---    & .220 \\
        $-1$  & .264   & .240   & .251   & .224 \\
        \hline
    \end{tabular}
\end{table}

\clearpage
\subsection*{Object segmentation: Matched DINOv3 re-run (SpelkeBench)}

\begin{table}[H]
    \centering
    \small
    \setlength{\tabcolsep}{4pt}
    \caption{\textbf{DINOv3 (Large) SpelkeBench IoU (layer $\times$ $\tau$).}
        DINOv3 run through the identical probe as ZWM (same cosine-NN rule, parameter-free LayerNorm, its own $\tau$ and layer sweep, 512px $\to$ $32\times32$ grid, no CRF). \textbf{bold} $=$ best per row; \texttt{final} $=$ post-LayerNorm.}
    \label{tab:dino_std}
    \begin{tabular}{lccccccccc}
        \\
        \hline
        Layer & 0.1 & 0.2 & 0.3 & 0.4 & 0.5 & 0.6 & 0.7 & 0.8 & 0.9 \\
        \hline
        L0    & .096 & .097 & .101 & .111 & .132 & .166 & .216 & \textbf{.245} & .220 \\
        L4    & .096 & .101 & .121 & .164 & .240 & .331 & \textbf{.335} & .215 & .084 \\
        L8    & .099 & .106 & .119 & .144 & .198 & .313 & \textbf{.421} & .247 & .064 \\
        L12   & .105 & .144 & .251 & .418 & \textbf{.437} & .307 & .171 & .075 & .034 \\
        L16   & .138 & .287 & .430 & \textbf{.435} & .327 & .208 & .113 & .055 & .032 \\
        L20   & .099 & .154 & .277 & .377 & \textbf{.396} & .333 & .219 & .113 & .047 \\
        L23   & .096 & .096 & .096 & .096 & .096 & .096 & .096 & .149 & \textbf{.547} \\
        final & .161 & .294 & .439 & .538 & \textbf{.568} & .538 & .462 & .336 & .157 \\
        \hline
    \end{tabular}
\end{table}

\begin{table}[H]
    \centering
    \small
    \setlength{\tabcolsep}{4pt}
    \caption{\textbf{Baby DINOv3 (Large) SpelkeBench IoU (layer $\times$ $\tau$).}
        Same matched probe as Table~\ref{tab:dino_std}. \textbf{bold} $=$ best per row; \texttt{final} $=$ post-LayerNorm.}
    \label{tab:dino_baby}
    \begin{tabular}{lccccccccc}
        \\
        \hline
        Layer & 0.1 & 0.2 & 0.3 & 0.4 & 0.5 & 0.6 & 0.7 & 0.8 & 0.9 \\
        \hline
        L0    & .212 & .242 & \textbf{.248} & .242 & .223 & .198 & .171 & .142 & .103 \\
        L4    & .179 & .266 & \textbf{.285} & .253 & .201 & .148 & .103 & .069 & .045 \\
        L8    & .148 & .272 & \textbf{.345} & .293 & .205 & .129 & .077 & .048 & .034 \\
        L12   & .142 & .331 & \textbf{.440} & .376 & .263 & .166 & .094 & .052 & .034 \\
        L16   & .148 & .337 & .482 & \textbf{.510} & .446 & .335 & .217 & .116 & .048 \\
        L20   & .118 & .235 & .394 & .495 & \textbf{.519} & .476 & .374 & .228 & .091 \\
        L23   & .109 & .173 & .291 & .418 & .497 & \textbf{.507} & .437 & .289 & .114 \\
        final & .154 & .264 & .387 & .482 & \textbf{.518} & .498 & .411 & .263 & .104 \\
        \hline
    \end{tabular}
\end{table}

For completeness, the same layer $\times$ $\tau$ sweep was run for the remaining representation baselines (Tables~\ref{tab:vjepa_std}--\ref{tab:resnet_sweep}), so that each of these models is reported at its own best configuration rather than at a setting inherited from another model.

\begin{table}[H]
    \centering
    \small
    \setlength{\tabcolsep}{4pt}
    \caption{\textbf{V-JEPA2 (Large) SpelkeBench IoU (layer $\times$ $\tau$).}
        Same matched probe as Table~\ref{tab:dino_std}, at 256px. \textbf{Bold} $=$ best per row; \texttt{final} $=$ post-LayerNorm. Cells at $\approx$0.096 are the whole-image floor, where the affinity threshold groups the entire image into one segment.}
    \label{tab:vjepa_std}
    \begin{tabular}{lccccccccc}
        \\
        \hline
        Layer & 0.1 & 0.2 & 0.3 & 0.4 & 0.5 & 0.6 & 0.7 & 0.8 & 0.9 \\
        \hline
        L0    & .096 & .096 & .097 & .098 & .102 & .129 & .184 & \textbf{.216} & .190 \\
        L4    & .096 & .096 & .096 & .096 & .096 & .096 & .096 & .096 & \textbf{.129} \\
        L8    & .096 & .096 & .096 & .096 & .096 & .096 & .096 & .096 & \textbf{.110} \\
        L12   & .096 & .096 & .096 & .096 & .096 & .096 & .096 & .096 & \textbf{.178} \\
        L16   & .096 & .096 & .096 & .096 & .096 & .096 & .096 & .107 & \textbf{.232} \\
        L20   & .096 & .096 & .096 & .096 & .096 & .103 & .203 & \textbf{.221} & .101 \\
        L23   & .099 & .128 & .209 & \textbf{.252} & .208 & .145 & .107 & .092 & .091 \\
        final & .099 & .128 & .209 & \textbf{.252} & .208 & .145 & .107 & .092 & .091 \\
        \hline
    \end{tabular}
\end{table}

\begin{table}[H]
    \centering
    \small
    \setlength{\tabcolsep}{4pt}
    \caption{\textbf{Baby V-JEPA2 (Large) SpelkeBench IoU (layer $\times$ $\tau$).}
        Same matched probe, at 256px. \textbf{Bold} $=$ best per row.}
    \label{tab:vjepa_baby}
    \begin{tabular}{lccccccccc}
        \\
        \hline
        Layer & 0.1 & 0.2 & 0.3 & 0.4 & 0.5 & 0.6 & 0.7 & 0.8 & 0.9 \\
        \hline
        L0    & .181 & .220 & .254 & \textbf{.274} & .268 & .250 & .224 & .194 & .157 \\
        L4    & .098 & .102 & .110 & .122 & .147 & .195 & .261 & \textbf{.268} & .175 \\
        L8    & .096 & .097 & .100 & .107 & .122 & .152 & .217 & \textbf{.298} & .199 \\
        L12   & .096 & .096 & .097 & .100 & .113 & .142 & .216 & \textbf{.318} & .191 \\
        L16   & .096 & .096 & .097 & .103 & .128 & .187 & .313 & \textbf{.323} & .150 \\
        L20   & .096 & .096 & .100 & .121 & .179 & .297 & \textbf{.372} & .268 & .125 \\
        L23   & .096 & .096 & .103 & .128 & .198 & .319 & \textbf{.373} & .255 & .120 \\
        final & .096 & .096 & .103 & .128 & .198 & .319 & \textbf{.373} & .255 & .120 \\
        \hline
    \end{tabular}
\end{table}

\begin{table}[H]
    \centering
    \small
    \setlength{\tabcolsep}{4pt}
    \caption{\textbf{ResNet50 (ImageNet-supervised) SpelkeBench IoU (stage $\times$ $\tau$).}
        Same matched probe, at 256px; rows are residual stages. \textbf{Bold} $=$ best per row.}
    \label{tab:resnet_sweep}
    \begin{tabular}{lccccccccc}
        \\
        \hline
        Stage & 0.1 & 0.2 & 0.3 & 0.4 & 0.5 & 0.6 & 0.7 & 0.8 & 0.9 \\
        \hline
        S0 & .096 & .096 & .096 & .098 & .103 & .114 & .136 & .179 & \textbf{.243} \\
        S1 & .096 & .096 & .096 & .096 & .097 & .108 & .148 & \textbf{.177} & .070 \\
        S2 & .096 & .096 & .099 & .121 & .214 & \textbf{.248} & .149 & .065 & .034 \\
        S3 & .096 & .112 & .235 & .388 & \textbf{.397} & .324 & .225 & .141 & .097 \\
        S4 & .234 & .300 & .331 & \textbf{.338} & .319 & .284 & .245 & .216 & .204 \\
        \hline
    \end{tabular}
\end{table}

\begin{table}[H]
    \centering
    \small
    \caption{\textbf{SpelkeBench segmentation, all methods (overall mIoU, 547 imgs).}
        res $=$ input resolution (patch grid for the feature-NN probes); feature-NN $=$ best over the layer $\times$ $\tau$ sweep. ZWM is evaluated at 256px throughout, so the 256 rows are the like-for-like comparison; the 512 rows give those baselines four times ZWM's input resolution. $\dagger$V-JEPA2 at 512 is additionally out of distribution for its processor, which normally fixes input to 256px. $\ddagger$V-JEPA2 (Giant) is run at its native 384px ($24\times24$), the same checkpoint used for the other evals.}
    \label{tab:seg_methods}
    \begin{tabular}{llcc}
        \\
        \hline
        Model & probe & res & mIoU \\
        \hline
        \multicolumn{4}{l}{\textit{ZWM: native zero-shot inference}} \\
        ZWM-170M (BigVideo)     & \textbf{native} (perturb-compare) & 256 & 0.522 \\
        ZWM-1B (BigVideo)       & \textbf{native} (perturb-compare) & 256 & \textbf{0.610} \\
        BabyZWM-170M (BabyView) & \textbf{native} (perturb-compare) & 256 & 0.496 \\
        BabyZWM-1B (BabyView)   & \textbf{native} (perturb-compare) & 256 & 0.500 \\
        \hline
        \multicolumn{4}{l}{\textit{feature-NN probe: 256px input}} \\
        ZWM-170M (BigVideo)     & feature-NN & 256 ($32\times32$) & 0.383 \\
        ZWM-1B (BigVideo)       & feature-NN & 256 ($32\times32$) & 0.385 \\
        BabyZWM-170M (BabyView) & feature-NN & 256 ($32\times32$) & 0.417 \\
        BabyZWM-1B (BabyView)   & feature-NN & 256 ($32\times32$) & 0.414 \\
        DINOv3 (Large)     & feature-NN & 256 ($16\times16$) & 0.484 \\
        Baby DINOv3 (Large)     & feature-NN & 256 ($16\times16$) & 0.484 \\
        V-JEPA2 (Large)    & feature-NN & 256 ($16\times16$) & 0.252 \\
        Baby V-JEPA2 (Large)    & feature-NN & 256 ($16\times16$) & 0.373 \\
        V-JEPA2 (Giant)    & feature-NN & 384 ($24\times24$)$\ddagger$ & 0.257 \\
        ResNet50    & feature-NN & 256 ($16\times16$) & 0.396 \\
        \hline
        \multicolumn{4}{l}{\textit{feature-NN probe: 512px input}} \\
        DINOv3 (Large)  & feature-NN & 512 ($32\times32$)            & 0.568 \\
        Baby DINOv3 (Large)  & feature-NN & 512 ($32\times32$)            & 0.519 \\
        V-JEPA2 (Large) & feature-NN & 512 ($32\times32$)$\dagger$   & 0.305 \\
        Baby V-JEPA2 (Large) & feature-NN & 512 ($32\times32$)$\dagger$   & 0.383 \\
        \hline
        \multicolumn{4}{l}{\textit{supervised (reference)}} \\
        SAM2                         & supervised & 256 & \textbf{0.622} \\
        Mask2Former-COCO-Panoptic    & supervised & 256 & 0.506 \\
        Mask2Former-COCO-Instance    & supervised & 256 & 0.362 \\
        Mask2Former-ADE-Panoptic     & supervised & 256 & 0.308 \\
        Mask2Former-Cityscapes-Pan.  & supervised & 256 & 0.078 \\
        Mask2Former-Cityscapes-Inst. & supervised & 256 & 0.032 \\
        \hline
    \end{tabular}
\end{table}

\clearpage
\subsection*{Ablation: training objective $\times$ readout mechanism}

To separate the contributions of the training objective and the readout mechanism, we cross the two dimensions while holding everything else fixed (same BabyZWM-170M architecture, same BabyView data, same capacity). ``Standard'' BabyZWM uses an asymmetric mask ($f_1$ fully visible, $f_2$ 90\%-masked), which prioritizes learning motion dynamics; we compare it against symmetric variants (45\%--45\% and 90\%--90\%) that resemble the masking used by popular masked video models such as VideoMAE~\cite{tong_videomae_2022}. We read each checkpoint two ways: with the matched, label-free feature-NN probe (which measures how decodable the frozen representation is) and with our native perturb-predict-compare mechanism (main-text Figure~\ref{fig:main_objective_readout}). Crossing the objective with the readout gives four primary settings (plus the 45\%--45\% variant, which is strictly worse on every task). Neither factor alone accounts for ZWM's performance: the full combination (asymmetric objective, causal-inference readout) is the best setting on every task, and degrading either factor costs performance on every task.

\begin{table}[H]
    \centering
    \small
    \caption{\textbf{Training objective $\times$ readout ablation (BabyZWM-170M).}
        Same architecture, BabyView data, and capacity; only the masking objective differs. Each checkpoint is read with the matched, label-free feature-NN probe (best over the layer sweep, best layer in parentheses) and with our native perturb-predict-compare mechanism. Optical flow: TAP-Vid-DAVIS position accuracy; relative depth: \% of pairs ordered correctly (chance $=$ 50\%); segmentation: SpelkeBench overall mIoU (547 images). Under the fixed probe, both symmetric variants show substantially degraded representations across all three tasks --- the representation itself is worse, not merely less amenable to the mechanism.}
    \label{tab:mask_ablation}
    \begin{tabular}{lcccccc}
        \\
        \hline
         & \multicolumn{2}{c}{optical flow} & \multicolumn{2}{c}{relative depth} & \multicolumn{2}{c}{segmentation} \\
        objective & feature-NN & native & feature-NN & native & feature-NN & native \\
        \hline
        asymmetric 90\% (standard) & 0.526 (L16) & 0.616 & 87.2 (L20) & 94.2 & 0.417 (L16) & 0.496 \\
        symmetric 90\%--90\%       & 0.344 (L16) & 0.542 & 69.5 (L16) & 71.8 & 0.379 (L12) & 0.479 \\
        symmetric 45\%--45\%       & 0.127 (L8)  & 0.482 & 54.3 (L12) & 73.8 & 0.163 (L0)  & 0.213 \\
        \hline
    \end{tabular}
\end{table}

\begin{table}[H]
    \centering
    \caption{\textbf{Feature-NN relative-depth per-layer sweep for the objective ablation (\% accuracy, chance $=$ 50\%).}
        The standard asymmetric objective climbs to 87.2\% at intermediate/deep layers, whereas both symmetric variants stay far lower at every layer --- the 45\%--45\% variant never exceeds $\sim$54\% (near chance). \textbf{Bold} $=$ best layer per model.}
    \label{tab:mask_depth_sweep}
    \begin{tabular}{lccc}
        \\
        \hline
        Layer & standard & symmetric 45\%--45\% & symmetric 90\%--90\% \\
        \hline
        L0    & 57.3          & 53.1          & 53.7 \\
        L4    & 72.0          & 48.1          & 54.9 \\
        L8    & 81.7          & 52.4          & 61.5 \\
        L12   & 84.8          & \textbf{54.3} & 64.6 \\
        L16   & 86.0          & 53.0          & \textbf{69.5} \\
        L20   & \textbf{87.2} & 53.0          & 59.8 \\
        final & 82.3          & 48.2          & 60.4 \\
        \hline
    \end{tabular}
\end{table}

Both readouts degrade together when the objective is weakened: the feature-NN probe does \emph{not} stay flat while only the native mechanism drops. Because the architecture and capacity are held fixed and only the objective changes, this isolates the temporally-factored, motion-prioritizing objective as what \emph{produces} the correspondence structure that both a fixed feature-matching probe and our zero-shot mechanism rely on. Mechanistically, predicting a heavily-masked second frame from a fully-visible first frame forces the model to learn cross-frame correspondence, which symmetric masking never requires; accordingly the most correspondence-dependent readouts (optical flow and binocular stereo depth) collapse hardest, with the 45\%--45\% variant falling to near chance on depth at every layer (Table~\ref{tab:mask_depth_sweep}).

\clearpage
\subsection*{Concentration of information in sparse patches}

Main-text Figure~\ref{fig:main_objective_readout}B asks how much of the target frame the predictor needs before it can reconstruct the rest. We sample 100 clips at random (fixed seed) from the Kinetics-400 validation set, skipping any clip shorter than 20 frames. From each we draw one frame pair: a uniformly random start index and a second frame $5$--$15$ frames later, comparable to the inter-frame gap used in training. Frames are resized and centre-cropped to $256 \times 256$ and patchified into $8 \times 8$ patches, giving $32 \times 32 = 1024$ patches per frame.

The context frame $f_1$ is fully visible throughout; only the proportion of $f_2$ revealed varies, over $\{0, 0.5, 1, 2, 3, 5, 7.5, 10, 15, 20, 30\}\%$ of the 1024 patches. Two details make the measurement non-trivial. First, before revealing anything we reserve a random 20\% of $f_2$'s patches (204 patches) as an evaluation set that is \emph{never} revealed at any proportion, and we score mean squared error \emph{only} on those patches; revealed patches are drawn exclusively from the remaining 820 patches. Without this reservation the error would fall simply because more of the frame is handed to the model verbatim rather than predicted, and the curve would measure the protocol rather than the representation. Second, the revealed sets are nested --- each larger set contains the smaller --- so adding a point only ever adds information.

We normalize each clip's curve to its own error with no patches revealed and report percentages; the mean is taken across clips and the 95\% confidence interval comes from 10{,}000 bootstrap resamples of the 100 clips, computed after normalization.

\begin{table}[H]
    \centering
    \caption{\textbf{Held-out prediction error versus the proportion of target-frame patches revealed.}
        BabyZWM-170M under the two training objectives, as \% of each clip's error with no patches revealed (mean over 100 Kinetics-400 clips, 95\% bootstrap CI in brackets). Error is scored only on a reserved 20\% of patches never revealed at any proportion.}
    \label{tab:patch_sufficiency}
    \begin{tabular}{lcc}
        \\
        \hline
        \% revealed & temporally factored (0--90) & symmetric (90--90) \\
        \hline
        0    & 100.0 & 100.0 \\
        0.5  & 48.9 [42.9, 55.0] & 59.5 [54.9, 64.0] \\
        1    & 35.1 [30.5, 39.8] & 44.7 [40.4, 48.9] \\
        2    & 25.3 [21.9, 28.7] & 34.0 [30.5, 37.4] \\
        3    & 19.9 [17.1, 22.8] & 28.4 [25.5, 31.4] \\
        5    & 15.8 [13.4, 18.2] & 23.6 [21.0, 26.3] \\
        7.5  & 12.6 [10.8, 14.5] & 20.1 [17.8, 22.5] \\
        10   & \textbf{10.5 [9.0, 12.0]} & \textbf{17.6 [15.5, 19.9]} \\
        15   & 8.0 [6.8, 9.2]   & 14.9 [12.9, 17.0] \\
        20   & 6.3 [5.4, 7.4]   & 13.1 [11.2, 15.1] \\
        30   & 4.9 [4.1, 5.8]   & 11.9 [10.1, 14.0] \\
        \hline
    \end{tabular}
\end{table}

Revealing 1\% of patches (10 of 1024) already reduces prediction error by 65\%, and 10\% reduces it by 90\%: the information needed to predict the frame is concentrated in a small subset of patches. ZWM's temporally-factored objective concentrates it more than the symmetric ablation, with lower error at every proportion.

\clearpage
\subsection*{Intuitive physics: Attention head analysis}

\begin{figure}[h]
    \centering
    \includegraphics[width=\textwidth]{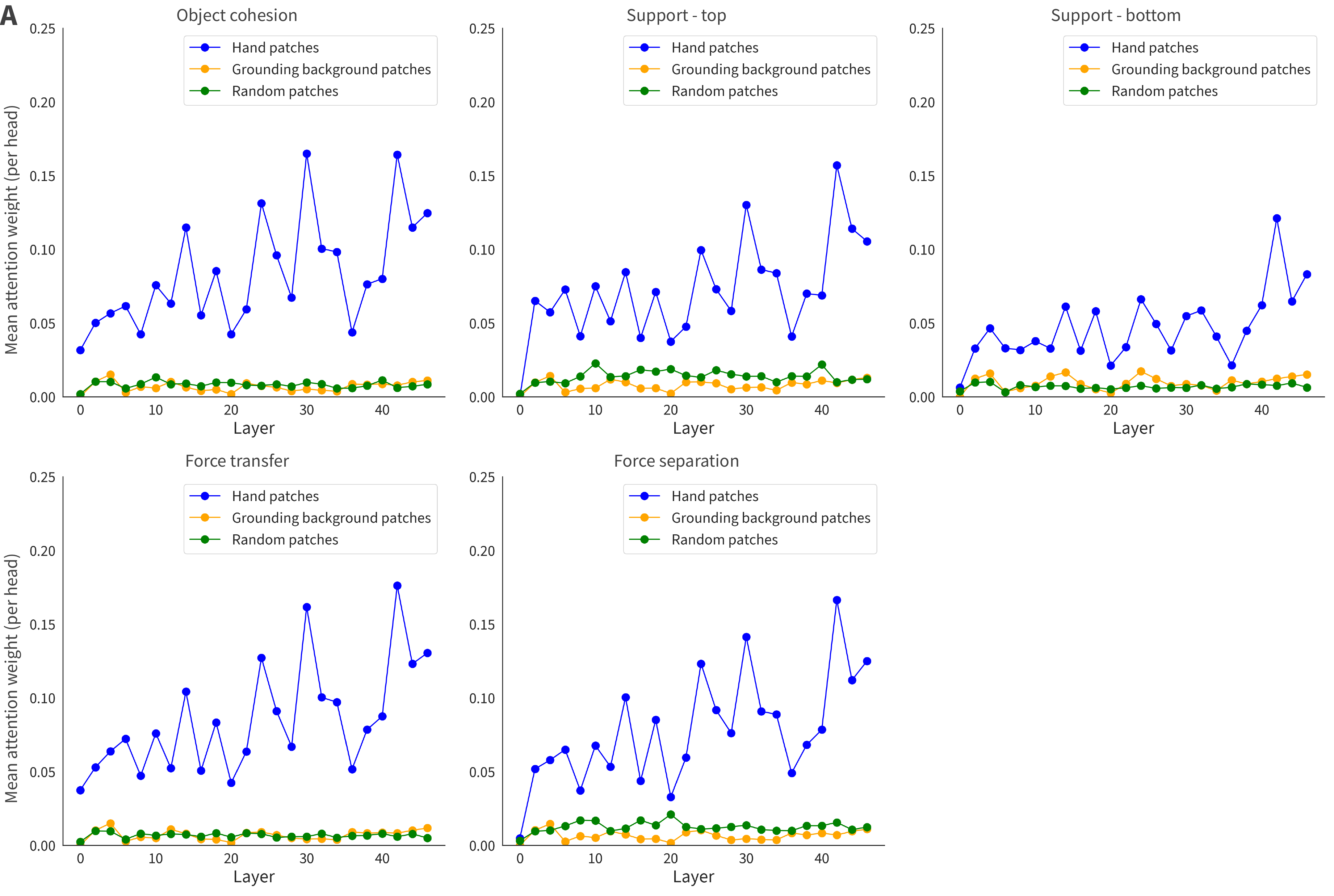}
    \caption{\textbf{Attention head analysis for intuitive physics.}
        Layer-wise average attention weights from the moved object's query patch to hand patches, background grounding patches, and random patches, shown for each intuitive physics category. In deeper layers, attention is disproportionately allocated to the hand---the causal agent of object motion---relative to background and random patches.}
    \label{fig:supp_attention_heads}
\end{figure}

To understand how BabyZWM implements intuitive physical reasoning internally, we analyze the attention patterns of individual transformer heads during the intuitive physics evaluation.

\paragraph{Methodology.}
For each intuitive physics example, we extract the full attention weight tensor across all layers and heads during the factual prediction forward pass. The model receives two frames as input: the context frame $f_1$ (1024 tokens) and the partially unmasked target frame $f_2$ (1024 tokens), yielding 2048 total tokens. We select a \textit{query patch} located on the moved object in $f_2$ (identified from human annotations of the object's position in the target frame) and examine which tokens this query patch attends to across all layers.

We partition the key tokens into three groups:
\begin{itemize}
    \item \textbf{Hand patches}: The 16 patches (forming a $4 \times 4$ patch region, i.e., $32 \times 32$ pixels) centered on the hand's revealed location in $f_2$---the causal agent responsible for the object's motion.
    \item \textbf{Background grounding patches}: The patches revealed at fixed background locations (top and bottom image borders) that anchor camera pose and illumination, providing non-causal contextual information.
    \item \textbf{Random patches}: An equal number of patches sampled randomly from $f_2$, excluding hand and background patches, serving as a baseline.
\end{itemize}

\paragraph{Quantitative analysis.}
For each layer, we compute the average attention weight (averaged across all heads) from the query patch to each of the three patch groups. To ensure comparability across groups of different sizes, we normalize the background attention by the ratio of background patches to hand patches. We average these layer-wise attention profiles across all examples within each intuitive physics category (e.g., object cohesion, support, force transfer) and across 8 random seeds.

\paragraph{Results.}
The layer-wise attention profiles reveal that in deeper transformer layers, attention from the moved object's query patch tends to be disproportionately allocated to the hand patches relative to both background grounding patches and random patches. This pattern is broadly consistent across intuitive physics categories, suggesting that the model may have learned to preferentially attend to the hand---the causal agent of object motion---when predicting physical outcomes. While these attention patterns are consistent with a ``causal attention'' interpretation, we note that attention weights are an indirect measure of the model's internal computations, and further mechanistic work would be needed to confirm whether these heads play a causal role in the model's physical predictions. Nonetheless, the emergence of hand-directed attention in deeper layers is suggestive of a hierarchical computation in which later layers begin to integrate information about agent--object relationships relevant to physical prediction. Layer-wise attention profiles for each intuitive physics category are shown in Figure~\ref{fig:supp_attention_heads}.

\clearpage
\subsection*{Intuitive physics: Flow-based motion-fidelity metric}

\begin{table}[H]
    \centering
    \caption{\textbf{Flow-based motion-fidelity metric for intuitive physics (BabyZWM-1B), per category.}
        ``Primary'' denotes the manipulated-object region. Acc $=$ \% of examples within 50\% normalized flow error; EPE in pixels (lower is better); graded $=$ fraction of the context$\to$target gap the prediction closes (0--1, higher is better). For force-separation, the secondary object has EPE 1.1\,px (correctly near-zero motion), so read EPE rather than \% accuracy there.}
    \label{tab:phys_motion_fidelity}
    \begin{tabular}{lcccc}
        \\
        \hline
        Category & Prim.\ acc & Prim.\ EPE (px) & Prim.\ graded & Overall acc \\
        \hline
        Cohesion         & 0.98 & 6.9  & 0.93 & 0.98 \\
        Support (top)    & 0.84 & 22.6 & 0.79 & 0.81 \\
        Support (bottom) & 0.83 & 21.3 & 0.74 & 0.83 \\
        Force transfer   & 0.93 & 14.4 & 0.89 & 0.86 \\
        Force separation & 0.99 & 9.2  & 0.90 & 0.99 \\
        \hline
        \textbf{Overall} & \textbf{0.91} & \textbf{14.9} & \textbf{0.85} & \textbf{0.90} \\
        \hline
    \end{tabular}
\end{table}

\begin{table}[H]
    \centering
    \caption{\textbf{Graded closeness score: model comparison across all categories.}
        Graded $=$ $d(\text{pred},f_1) / [d(\text{pred},f_1) + d(\text{pred},f_2)] \in [0,1]$ (1 $=$ matches the correct outcome $f_2$; 0 $=$ predicts no change, i.e.\ stays at the context $f_1$). MSE and flow graded require a pixel prediction and exist only for the generative ZWM models; feature graded (cosine distance) is the only version available for V-JEPA2, which outputs no pixels. Accuracy (all categories combined) is whole-image accuracy---whether the prediction is closer to the correct outcome ($f_2$) than to the context ($f_1$)---and is reported under each distance that applies to a given model class: pixel MSE and LPIPS for the generative ZWM models, and feature cosine for V-JEPA2, which outputs no pixels (the same whole-image metric used for V-JEPA2 in the main-text object-reasoning figure). MSE and LPIPS capture different things, raw pixel error versus learned perceptual similarity, yet agree to within 0.04 on every ZWM model, so the result does not hinge on the choice of distance. Accuracy is near-ceiling for the stronger models; the graded columns, on the primary (manipulated-object) region, provide the finer comparison. V-JEPA2's feature graded sits near 0.5 because cosine distances between frames are compressed, so it is not on the same scale as the ZWM pixel and flow graded.}
    \label{tab:phys_graded}
    \begin{tabular}{lcccccc}
        \\
        \hline
        & \multicolumn{3}{c}{accuracy} & \multicolumn{3}{c}{graded} \\
        \cmidrule(lr){2-4}\cmidrule(lr){5-7}
        Model & MSE & LPIPS & feature & MSE & flow & feature \\
        \hline
        ZWM-170M (BigVideo) & 0.98 & 1.00 & --- & 0.87 & 0.90 & --- \\
        ZWM-1B (BigVideo)   & 1.00 & 1.00 & --- & 0.90 & 0.93 & --- \\
        BabyZWM-170M    & 0.94 & 0.98 & --- & 0.77 & 0.79 & --- \\
        BabyZWM-1B      & 0.95 & 0.99 & --- & 0.84 & 0.85 & --- \\
        V-JEPA2 (Large) & ---  & ---  & 0.96 & ---  & ---  & 0.57 \\
        V-JEPA2 (Giant) & ---  & ---  & 0.91 & ---  & ---  & 0.62 \\
        Baby V-JEPA2 (Large) & --- & --- & 0.37 & ---  & ---  & 0.43 \\
        \hline
    \end{tabular}
\end{table}

\clearpage
\subsection*{Intuitive physics: Qualitative examples}

\begin{figure}[H]
    \centering
    \includegraphics[width=0.85\textwidth]{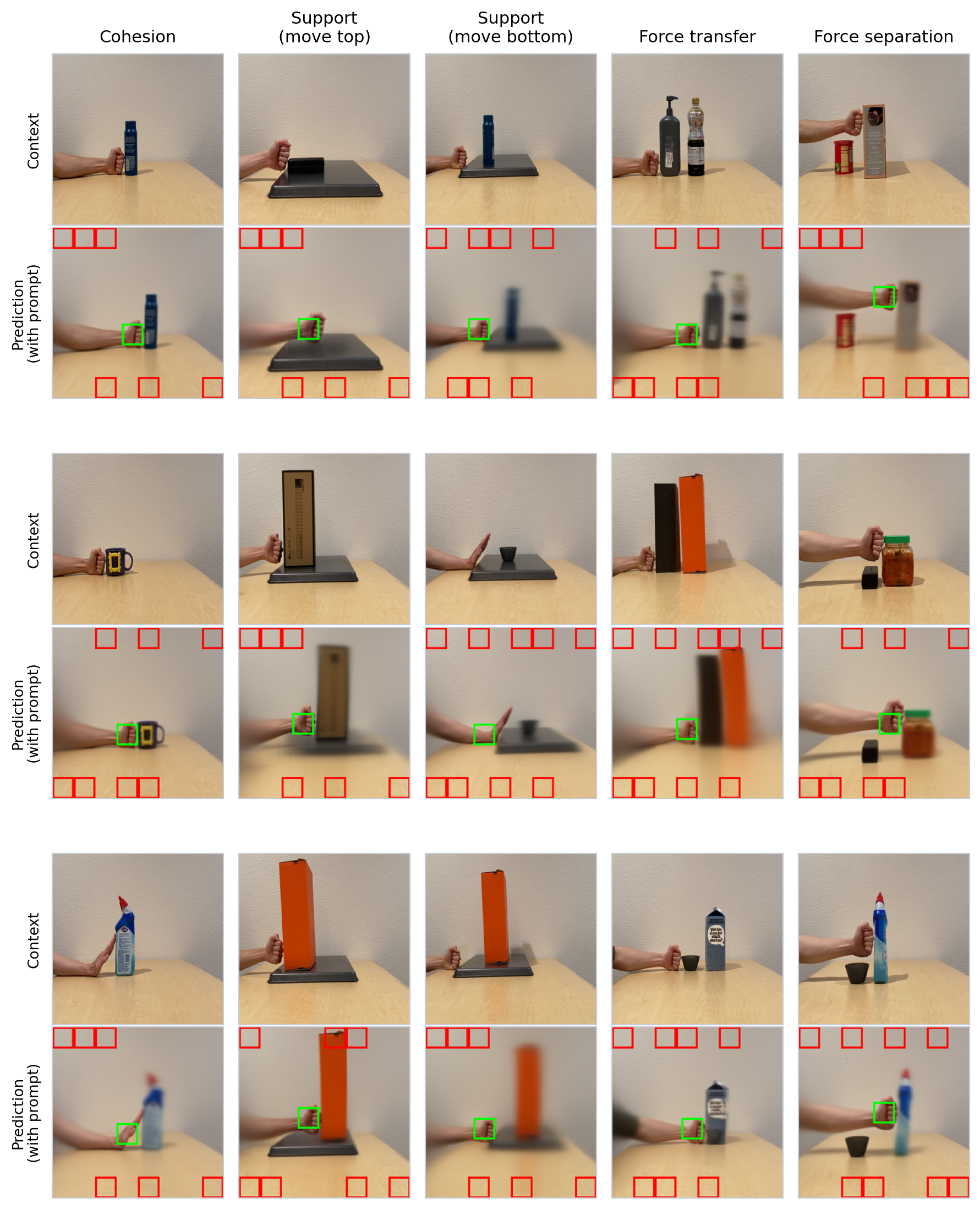}
    \caption{\textbf{Qualitative intuitive-physics predictions (BabyZWM-1B).}
        Three randomly selected examples per category; each pair shows the context frame
        (top) above the model's prediction for the masked target frame (bottom). Green marks
        the revealed patch giving the hand's target location, red the fixed border patches
        signalling a static camera --- these are the only target-frame patches the model receives.}
    \label{fig:supp_physics_qualitative}
\end{figure}

Support predictions are visibly less sharp than cohesion and force separation (Figure~\ref{fig:supp_physics_qualitative}), consistent with the graded motion-fidelity scores in Table~\ref{tab:phys_motion_fidelity}.

\clearpage
\subsection*{Intuitive physics: Mass sensitivity}

\begin{figure}[h]
    \centering
    \includegraphics[width=\textwidth]{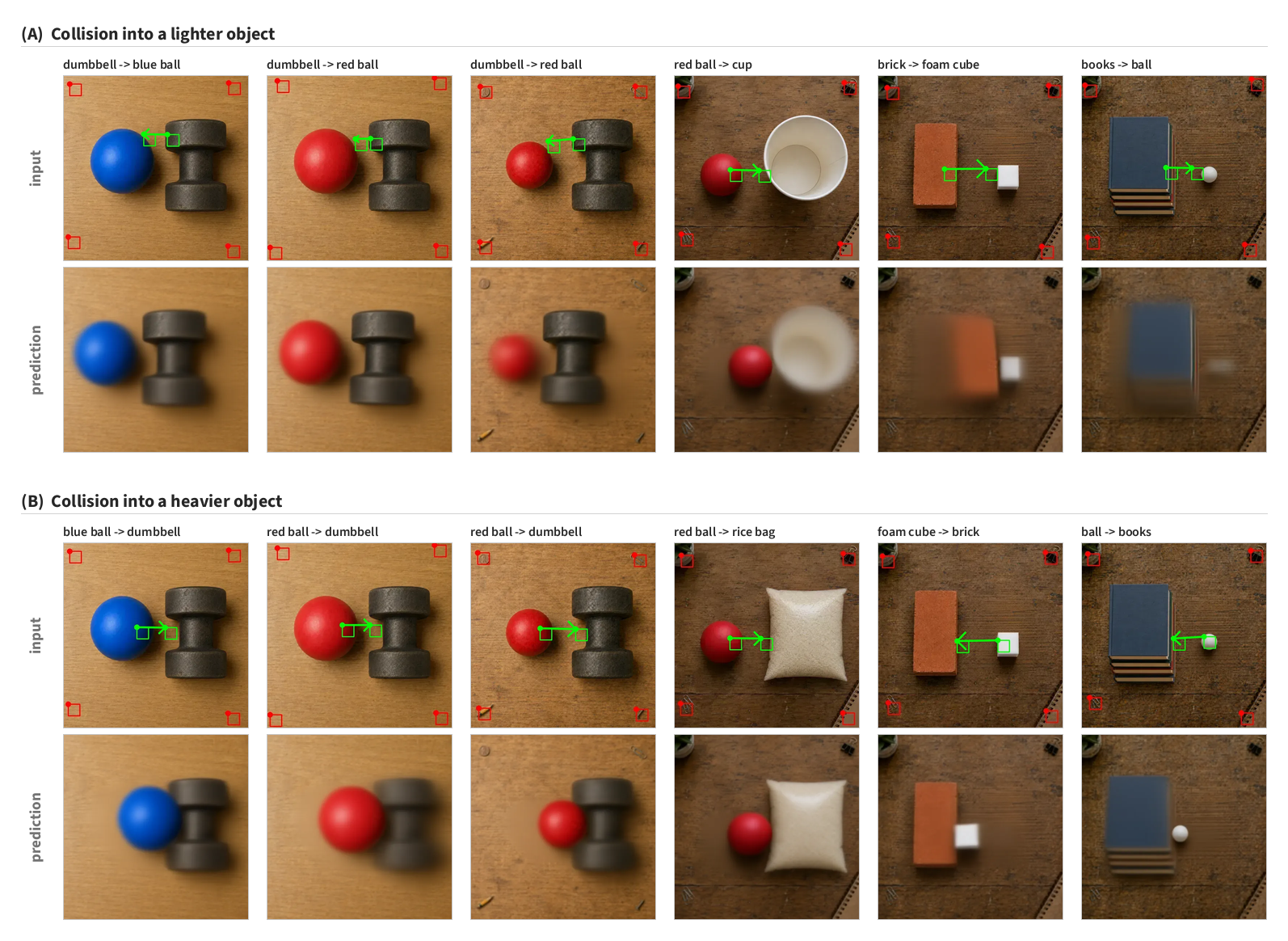}
    \caption{\textbf{Preliminary qualitative test of mass sensitivity (BabyZWM-1B).}
        Each column is one scene: the annotated input above, the model's prediction below.
        Green marks the prompted motion (the start and end position of the moved patch); red marks
        the background patches that hold the scene fixed. No human hand appears in any scene.
        \textbf{(A)} The heavier object is moved into the lighter one, which the model predicts to
        be additionally displaced by the contact. \textbf{(B)} The lighter object is moved into the
        heavier one, which the model predicts to stay in place while the lighter object comes to
        rest against it.}
    \label{fig:supp_mass_sensitivity}
\end{figure}

Our main intuitive-physics benchmark is not designed to test mass sensitivity. There, a human hand
drives the motion, and a hand-pushed light object is still physically expected to cause a heavier
one to move---so the two directions carry the same expected outcome and no asymmetry is available to
measure. We therefore ran a separate preliminary qualitative test in which no hand appears, so that
the only thing that can displace the struck object is the object that strikes it.

\paragraph{Stimuli.}
We use tabletop scenes containing a heavier and a lighter object, whose masses are evident from
familiar appearance. Heavy objects include a dumbbell, a brick, a stack of books, and a bag of rice;
light objects include a rubber ball, a foam cube, and an empty paper cup. We hold the scene fixed
using background patches, as in the main intuitive-physics evaluation, and vary whether we move the
lighter object into the heavier object, or vice versa. Notably, we do not use a human hand to drive
the motion, for the reason above. In most scenes we hold the scene itself fixed---the same two
objects, positions, and contact geometry---and reverse only which of the two is moved; in one further
contrast we instead hold the moved object fixed (the same ball, the same motion) and vary only the
object it is moved into, from a light paper cup to a heavy bag of rice. This gives 12 scenes in
total, all run through BabyZWM-1B.

\paragraph{Results.}
The predictions are mass-sensitive (Figure~\ref{fig:supp_mass_sensitivity}). When moving a lighter
object into a heavier one, the model predicts the lighter object to travel and come to rest against
the heavier one, which stays in place. Whereas when moving the heavier object into the lighter one,
the model predicts the lighter object to be additionally displaced by the contact. 
However, these results are preliminary and qualitative: the sample is small and run at a single seed, and we
do not report a quantitative mass-sensitivity metric or vary the mass ratio parametrically.

\paragraph{An in-principle argument.}
Mass is not a special or separate property of the world requiring dedicated machinery. It is simply another regularity in how objects behave on contact, alongside support, cohesion, and force transfer --- all of which BabyZWM does acquire from BabyView. A child's everyday visual experience contains many interactions between objects of differing apparent mass, so a model that learns those other regularities from naturalistic video faces no principled barrier to also learning that heavier-looking objects resist being displaced.

\paragraph{Why a rigorous test needs a richer interface.}
Testing this rigorously calls for a richer interface than our current ZWM model. These tests carry inherent ambiguity: how fast or forcefully the moving object was travelling (our prompt specifies its displacement, not its speed), and how heavy either object is, are not observable from the input --- and therefore neither is how likely the struck object is to move, nor how far. ZWM makes a single deterministic prediction, so it cannot express such a graded expectation; and because it is trained with a regression objective, its prediction under ambiguity averages over the possible outcomes, which shows up as blurry predictions. In follow-up work we are therefore building a variant of ZWM that is distributional; its readout exposes the predicted distribution over how likely the struck object is to move, and how far. It is also multi-frame and has time embeddings, allowing us to specify the speed of object motion. We plan to use it to test mass sensitivity and other physical concepts more rigorously and quantitatively.

\clearpage
\subsection*{Developmental trajectory: emergence timing}

\begin{table}[H]
    \centering
    \caption{\textbf{Developmental emergence timing (BabyZWM-170M).}
        Emergence is measured over each capability's learnable gain (from a baseline to its max across checkpoints). \% training $=$ (training step $/$ 200{,}000) $\times$ 100, and the equivalent days of waking experience $=$ \% training $\times$ 0.95 (a full run $\approx$ 95 days); lower $=$ emerges earlier. 95\% CIs in brackets (bootstrap).}
    \label{tab:emergence_timing}
    \begin{tabular}{lccl}
        \\
        \hline
        Capability & \% training $\to$ 50\% gain & \% training $\to$ 80\% gain & Final-third trend \\
        \hline
        Relative depth      & 3.3 [3.0, 3.8]   & 4.6 [4.2, 15.1]  & fluctuates near ceiling \\
        Optical flow        & 3.6 [3.1, 4.5]   & 17.4 [5.3, 63.6] & rising slowly \\
        Intuitive physics   & 5.4 [5.3, 5.7]   & 8.7 [8.4, 9.0]   & peaked early \\
        Object segmentation & 12.6 [9.2, 20.0] & 69.8 (lower bd)  & \textbf{rising} (+40\%/100k, CI 28--52) \\
        \hline
    \end{tabular}
\end{table}

\clearpage
\subsection*{Neural alignment: predictivity results}

\begin{figure}[h]
    \centering
    \includegraphics[width=\textwidth]{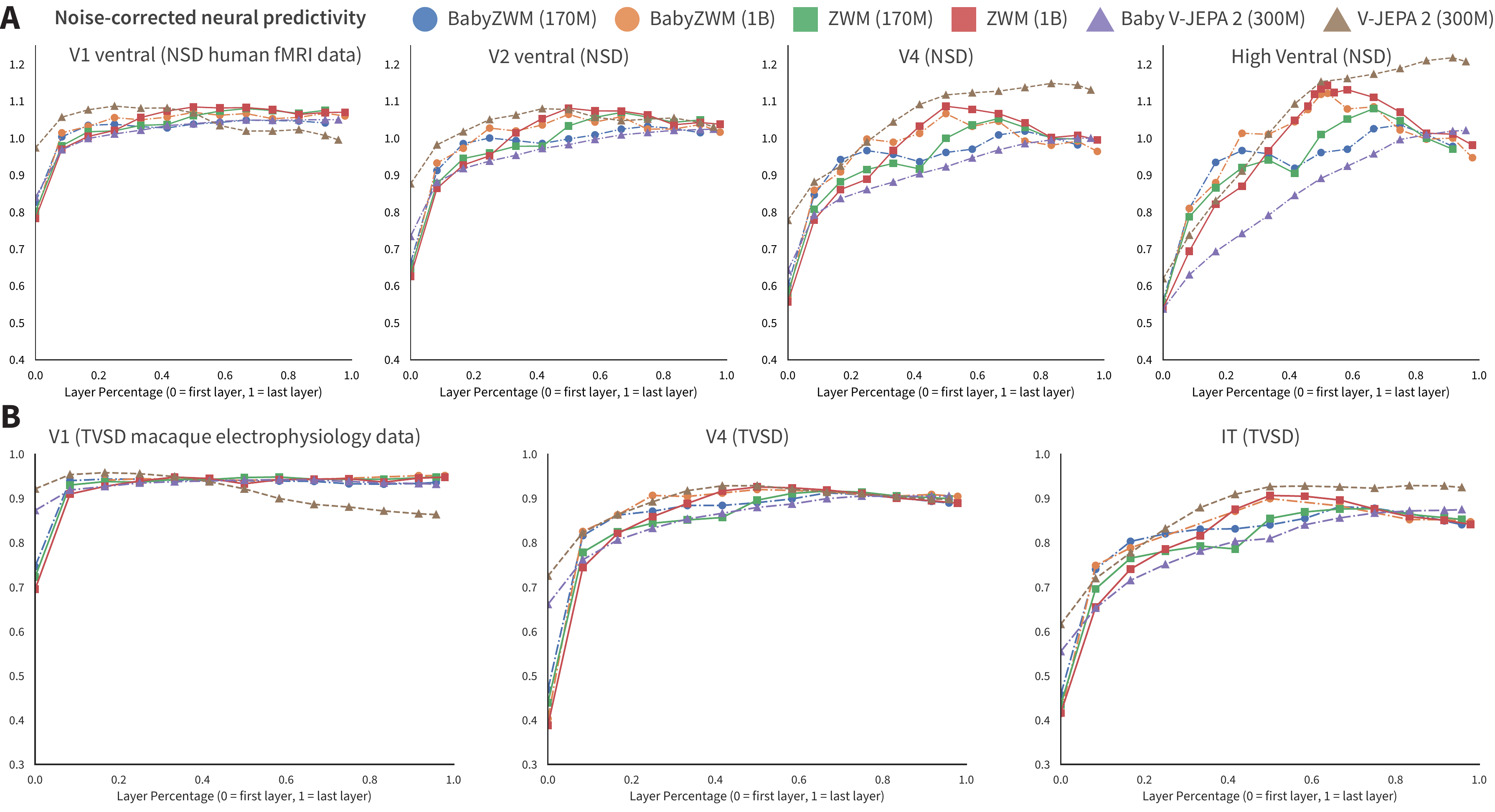}
    \caption{\textbf{Neural predictivity across NSD and TVSD.}
        (\textbf{A})~Cross-validated noise-corrected correlations between model features and human fMRI responses (NSD). (\textbf{B})~Cross-validated noise-corrected correlations between model features and macaque single-neuron electrophysiology responses (TVSD). Both benchmarks show hierarchically organized layer-area correspondences for BabyZWM and baseline models.}
    \label{fig:supp_neural_predictivity}
\end{figure}

In the main text, we reported that BabyZWM's internal representations align with hierarchical visual cortex organization using the Natural Scenes Dataset (NSD)~\cite{allen_massive_2022}. Here we present expanded neural predictivity results across both NSD (human fMRI) and the THINGS Ventral Stream Spiking Dataset (TVSD; macaque single-neuron electrophysiology)~\cite{papale_extensive_2025}, providing converging evidence across species and measurement modalities.

Across both benchmarks, BabyZWM exhibits hierarchically organized layer-area correspondence: earlier model layers best predict earlier cortical regions, while deeper layers align with higher visual areas. In NSD, this pattern is evident across all evaluated ROIs; TVSD corroborates these findings at single-neuron resolution, confirming that the representational hierarchy is not an artifact of the fMRI measurement scale. BabyZWM achieves predictivity comparable to ZWM variants trained on substantially larger and more diverse datasets, whereas Baby V-JEPA2 shows lower neural alignment than its larger-data counterpart. Full layer-by-region predictivity profiles for both benchmarks are shown in Figure~\ref{fig:supp_neural_predictivity}.

\begin{figure}[h]
    \centering
    \includegraphics[width=\textwidth]{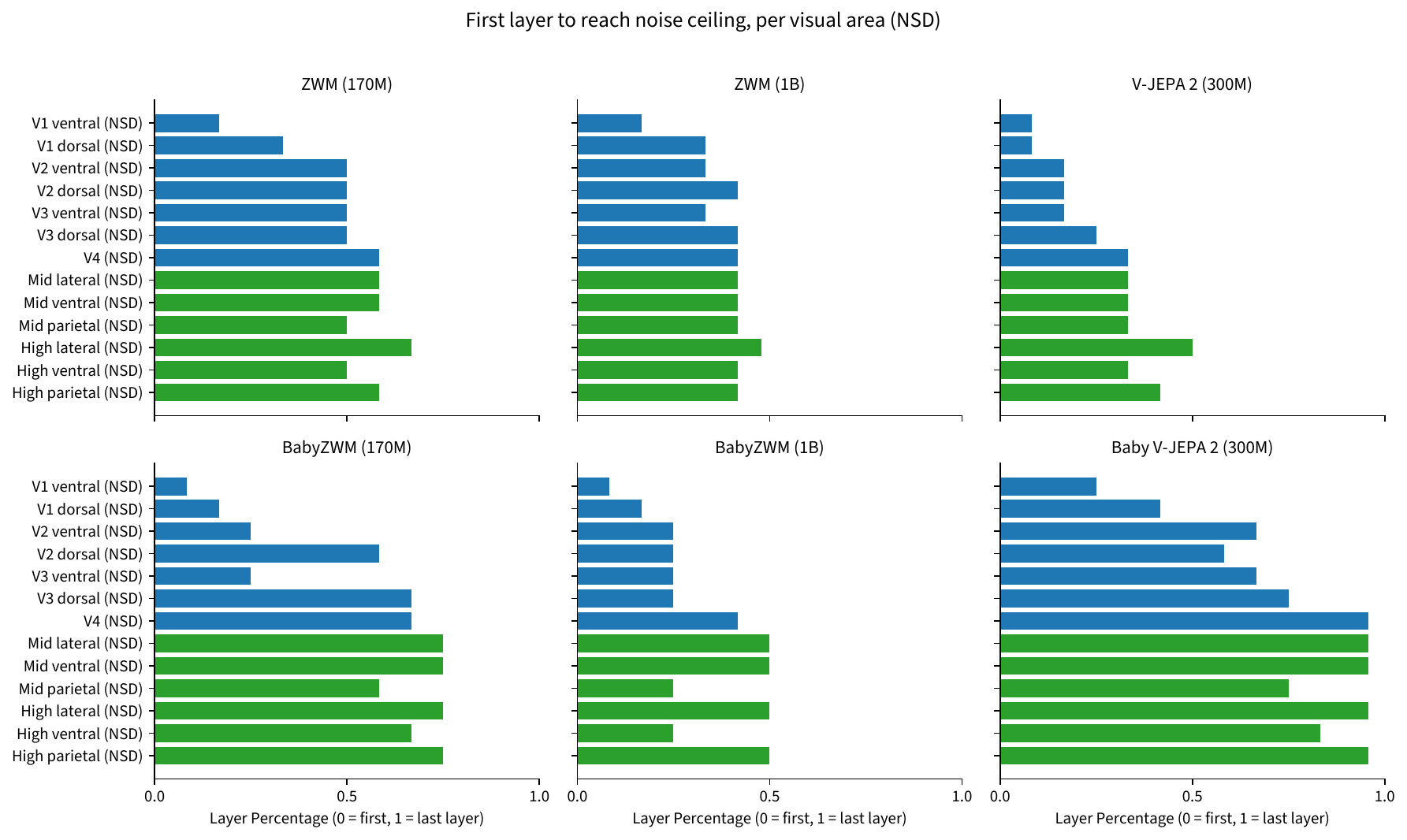}
    \caption{\textbf{Neural predictivity: layer-area correspondence across models (NSD).}
        One panel per model, arranged so each column pairs an architecture (170M, 1B, V-JEPA2) with its BabyView-trained counterpart (top: standard data; bottom: developmental diet). For each NSD visual area (rows, early areas in blue at top to mid/high areas in green at bottom), the bar shows the fractional encoder depth ($0=$ first layer, $1=$ last layer) of the earliest layer to reach the noise ceiling (else the best layer); a profile whose bars lengthen from early (blue) to higher (green) areas indicates hierarchical layer-area correspondence, with earlier model layers best predicting earlier cortical regions and deeper layers higher regions. Every model shows this hierarchy, so BabyZWM's correspondence --- from a single child's video --- matches the much-larger-data baselines. Across the developmental diet (each column, top vs.\ bottom), ZWM's correspondence is largely unchanged whereas V-JEPA2's shifts to its late layers, complementing the baby-data penalty (Table~\ref{tab:baby_data_penalty}). The readout is a coarse, ordinal summary, so the rising trend and the standard-vs-baby shift should be read rather than exact bar heights.}
    \label{fig:supp_layer_area_allmodels}
\end{figure}

To read this layer-area correspondence relative to alternatives, Figure~\ref{fig:supp_layer_area_allmodels} repeats the main-text Figure~\ref{fig:main_brain}C analysis for every model: per NSD visual area, the fractional encoder depth of the earliest layer to reach the noise ceiling (else the best layer). Two patterns hold. First, hierarchical layer-area correspondence is present in every model --- earlier layers reach the ceiling for earlier visual areas and progressively deeper layers for higher areas --- so BabyZWM's hierarchy, learned from a single child's video, is neither an artifact nor unique to it, and is comparable to models trained on orders of magnitude more data, consistent with established accounts of hierarchical visual representation. Second, the developmental diet affects the two objectives differently (each column pairs a standard model, top, with its BabyView-trained counterpart, bottom): ZWM's correspondence is essentially unchanged under the baby diet, whereas V-JEPA2's shifts markedly deeper, reaching the ceiling only in its late layers even for early areas. This complements the baby-data penalty (Table~\ref{tab:baby_data_penalty}) --- feature-prediction degrades under a child's data budget while the ZWM objective remains robust --- here visible as \emph{where in depth} alignment is achieved rather than its magnitude. Because the readout marks the first layer to reach the noise ceiling (else the best layer), it is a coarse, ordinal summary; we therefore read the rising trend and the standard-vs-baby shift, not exact bar heights.

\paragraph{An objective-based account of the layer-wise profiles.}
As a provisional hypothesis, we offer an objective-based account of the layer-wise shapes in main-text Figure~\ref{fig:main_brain}D: the profiles separate cleanly by training target, with ZWM peaking at intermediate depth and then declining toward its final layers (most so for higher areas), whereas V-JEPA2 rises monotonically to a late peak. We interpret this as a consequence of what each model predicts. ZWM's terminal layers are optimized to predict pixels, an appearance-level target not aligned with high-level cortex, so its deepest representations specialize away from IT- and high-ventral-like features; V-JEPA2 predicts high-level latent features and so keeps its deepest layers aligned with higher areas.

\paragraph{Robustness to a developmental data diet.}
To read BabyZWM's alignment relative to alternatives, we quantify the \emph{baby-data penalty}: the change in peak (best-layer) noise-corrected predictivity when a fixed architecture is retrained on BabyView rather than its standard training set, which isolates the effect of the data regime from the architecture and objective (Table~\ref{tab:baby_data_penalty}). Averaged over higher-order visual areas, V-JEPA2 loses $10.4\%$ of its peak predictivity, whereas ZWM loses only $2.1\%$ (170M) and $1.5\%$ (1B) --- a roughly five-fold smaller penalty. Moreover, the V-JEPA2 penalty grows monotonically along the visual hierarchy (from $-3.4\%$ at V1 ventral to $-16.2\%$ at high ventral), most damaging exactly the higher-area representations of greatest interest, while ZWM's penalty stays small and roughly flat across areas and species. This pattern replicates across both NSD and TVSD, indicating that the ZWM objective is more robust to a developmental data diet than feature-prediction.

\begin{table}[H]
    \centering
    \caption{\textbf{Baby-data penalty: robustness of neural predictivity to a developmental data diet.}
        Percentage change in peak (best-layer) noise-corrected predictivity when a fixed architecture is retrained on BabyView rather than its standard training set (negative $=$ drop under the developmental diet). Comparisons are architecture-matched: V-JEPA2 vs.\ Baby V-JEPA2, and ZWM vs.\ BabyZWM at 170M and 1B. The final row averages over higher-order visual areas (NSD V4, mid-ventral, and high-ventral; macaque V4 and IT).}
    \label{tab:baby_data_penalty}
    \begin{tabular}{lccc}
        \\
        \hline
        Region & V-JEPA2 (300M) & ZWM (170M) & ZWM (1B) \\
        \hline
        V1 ventral (NSD)   & $-3.4$  & $-2.9$ & $-1.3$ \\
        V2 ventral (NSD)   & $-5.1$  & $-3.5$ & $-1.5$ \\
        V3 ventral (NSD)   & $-8.3$  & $-3.3$ & $-1.5$ \\
        V4 (NSD)           & $-12.9$ & $-3.2$ & $-1.8$ \\
        Mid ventral (NSD)  & $-14.6$ & $-3.3$ & $-2.3$ \\
        High ventral (NSD) & $-16.2$ & $-4.1$ & $-1.9$ \\
        V1 (TVSD)          & $-1.7$  & $-0.5$ & $+0.4$ \\
        V4 (TVSD)          & $-2.5$  & $-0.4$ & $-0.6$ \\
        IT (TVSD)          & $-5.8$  & $+0.6$ & $-0.8$ \\
        \hline
        Higher-area mean   & $\mathbf{-10.4}$ & $\mathbf{-2.1}$ & $\mathbf{-1.5}$ \\
        \hline
    \end{tabular}
\end{table}

\begin{figure}[h]
    \centering
    \includegraphics[width=\textwidth]{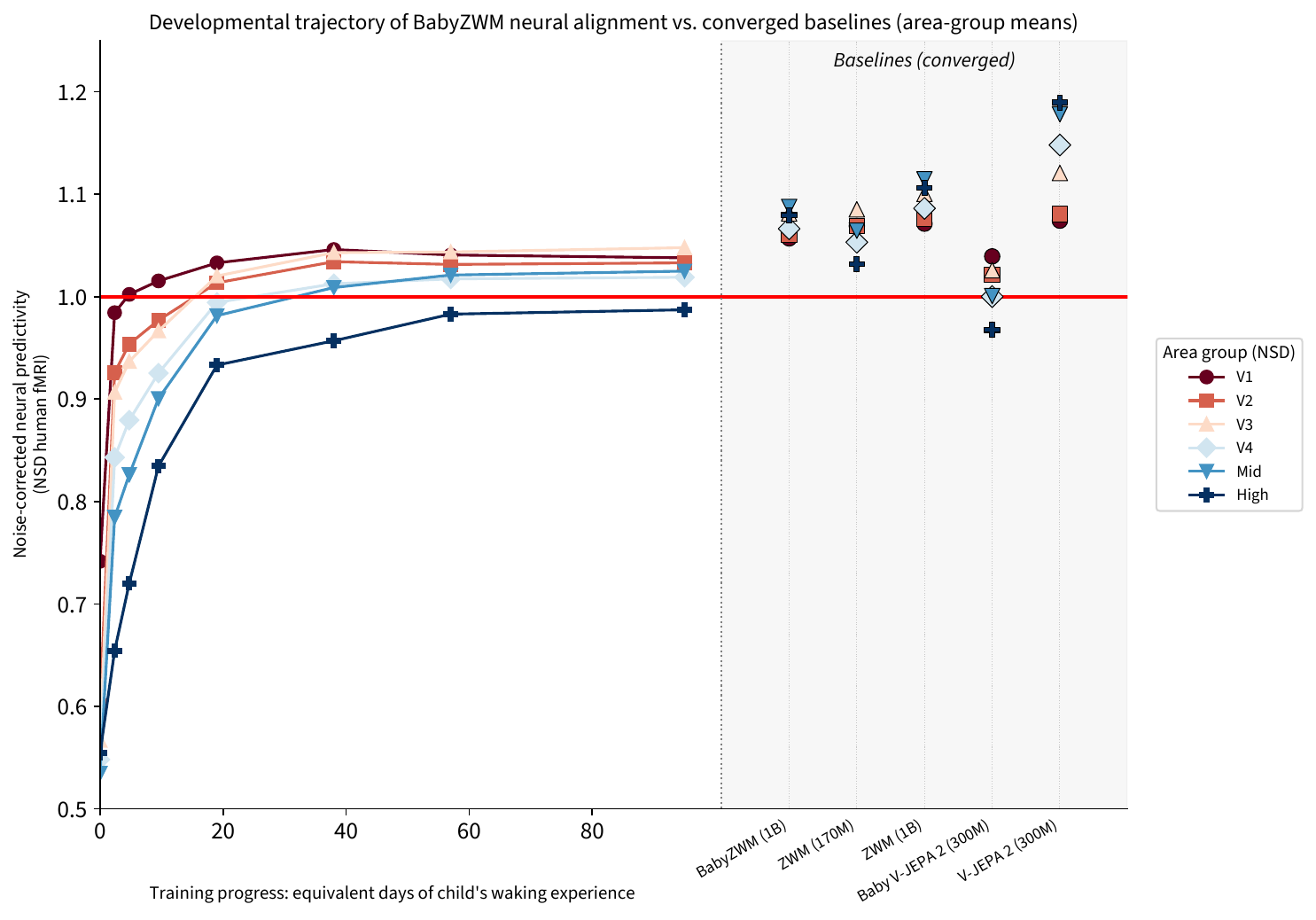}
    \caption{\textbf{Neural predictivity: developmental trajectory of BabyZWM relative to converged baselines (NSD).}
        BabyZWM-170M's noise-corrected predictivity is plotted against training progress (equivalent days of a child's waking experience), summarized as the mean over NSD ROIs within each cortical area group (V1--High); the bold red line marks the noise ceiling (predictivity $=1$). To the right of the divider, each baseline's converged (best-layer) predictivity is shown as endpoint markers on the same axis; the baselines have no comparable training-progress axis (it is calibrated to BabyZWM's developmental diet), so they are shown as converged endpoints rather than trajectories. Two comparisons matter. (i)~Under a matched developmental diet, BabyZWM aligns better than Baby V-JEPA2 at every area, and the gap widens up the hierarchy (higher-area mean predictivity $0.97$ vs.\ $0.96$ at 170M, and $1.01$ vs.\ $0.96$ at 1B). (ii)~The off-the-shelf V-JEPA2 is the single strongest higher-area predictor, but this advantage is bought entirely by large-scale, non-developmental training data: retraining the same architecture on BabyView (Baby V-JEPA2) collapses it to the weakest model in higher areas, below BabyZWM. Predictivity above the noise ceiling falls in a regime where the noise-corrected ratio is dominated by ceiling-estimation noise and should be read cautiously. Together these support reading BabyZWM's brain alignment as strong relative to developmentally-matched alternatives.}
    \label{fig:supp_dev_trajectory_baselines}
\end{figure}

\paragraph{Developmental trajectory relative to converged baselines.}
The main-text developmental trajectory (Figure~\ref{fig:main_brain}B) plots predictivity against training progress for BabyZWM alone, because its x-axis --- equivalent days of a child's waking experience --- is calibrated to BabyZWM's developmental diet and is not shared by models trained on different data and schedules; a baseline therefore cannot be placed on it as a trajectory. To still read BabyZWM's alignment relative to alternatives, Figure~\ref{fig:supp_dev_trajectory_baselines} keeps BabyZWM's trajectory and adds each baseline's converged (best-layer) predictivity as endpoint markers in a horizontally extended margin, on the same predictivity axis. The comparison reinforces the baby-data-penalty result: under a matched developmental diet, BabyZWM aligns better with cortex than Baby V-JEPA2 at every area, and although the off-the-shelf V-JEPA2 is the single strongest higher-area predictor, that advantage is bought entirely by its large-scale, non-developmental training data.

\begin{table}[H]
    \centering
    \small
    \caption{\textbf{Converged neural predictivity across models and cortical areas.}
        Peak (best-layer) noise-corrected predictivity at convergence, for each model and cortical area; NSD values are averaged over the ROIs within each area group, TVSD values are single ROIs. The final column averages over higher-order visual areas (NSD V4, mid-ventral, high-ventral; macaque V4 and IT), matching the baby-data-penalty definition (Table~\ref{tab:baby_data_penalty}). These are the converged endpoints plotted in Figure~\ref{fig:supp_dev_trajectory_baselines}. Under a matched developmental diet, BabyZWM aligns better than Baby V-JEPA2 at every area, and the gap widens up the hierarchy. Off-the-shelf V-JEPA2 is the strongest higher-area predictor, but retraining the same architecture on the developmental BabyView diet (Baby V-JEPA2) collapses it to the weakest model in higher areas. Values above $1.0$ exceed the noise ceiling and should be read cautiously, as the noise-corrected ratio there is increasingly dominated by ceiling-estimation noise.}
    \label{tab:converged_predictivity}
    \begin{tabular}{lcccccccccc}
        \\
        \hline
        & \multicolumn{6}{c}{NSD (human fMRI)} & \multicolumn{3}{c}{TVSD (macaque)} & Higher-area \\
        \cmidrule(lr){2-7}\cmidrule(lr){8-10}
        Model & V1 & V2 & V3 & V4 & Mid & High & V1 & V4 & IT & mean \\
        \hline
        BabyZWM (170M)         & 1.04 & 1.03 & 1.05 & 1.02 & 1.02 & 0.99 & 0.94 & 0.91 & 0.88 & 0.97 \\
        BabyZWM (1B)           & 1.06 & 1.06 & 1.08 & 1.07 & 1.09 & 1.08 & 0.95 & 0.92 & 0.90 & 1.01 \\
        ZWM (170M)             & 1.07 & 1.07 & 1.09 & 1.05 & 1.06 & 1.03 & 0.95 & 0.92 & 0.88 & 0.99 \\
        ZWM (1B)               & 1.07 & 1.08 & 1.10 & 1.09 & 1.11 & 1.11 & 0.95 & 0.93 & 0.91 & 1.03 \\
        Baby V-JEPA2 (300M)    & 1.04 & 1.02 & 1.03 & 1.00 & 1.00 & 0.97 & 0.94 & 0.91 & 0.87 & 0.96 \\
        V-JEPA2 (300M)         & 1.07 & 1.08 & 1.12 & 1.15 & 1.18 & 1.19 & 0.96 & 0.93 & 0.93 & 1.07 \\
        \hline
    \end{tabular}
\end{table}

\newpage




\clearpage



\end{document}